\documentclass[manuscript,screen,natbib]{acmart}

\usepackage{acronym}
\usepackage[skip=0pt]{caption}
\usepackage[inline]{enumitem}
\usepackage{booktabs}
\usepackage{makecell}
\usepackage{multirow}
\usepackage{subcaption}
\usepackage{xcolor}
\usepackage{colortbl}
\usepackage{fontawesome}
\usepackage{array}
\usepackage{listings}
\AtBeginDocument{%
  \providecommand\BibTeX{{%
    \normalfont B\kern-0.5em{\scshape i\kern-0.25em b}\kern-0.8em\TeX}}}

\newcommand{\header}[1]{\vspace{1mm}\noindent\textbf{#1.}}

\acrodef{CS}{conversational search}
\acrodef{IR}{information retrieval}

\allowdisplaybreaks

\copyrightyear{2025}
\acmYear{2025}
\setcopyright{cc}
\setcctype{by-sa}
\acmConference[CHIIR '25]{2025 ACM SIGIR Conference on Human Information Interaction and Retrieval}{March 24--28, 2025}{Melbourne, VIC, Australia}
\acmBooktitle{2025 ACM SIGIR Conference on Human Information Interaction and Retrieval (CHIIR '25), March 24--28, 2025, Melbourne, VIC, Australia}
\acmDOI{10.1145/3698204.3716464}
\acmISBN{979-8-4007-1290-6/2025/03}

\author{Clemencia Siro}
\authornote{Equal contributions.}
\orcid{0000-0001-5301-4244}
\affiliation{%
  \institution{University of Amsterdam}
  \city{Amsterdam}
  \country{The Netherlands}}
\email{c.n.siro@uva.nl}

\author{Zahra Abbasiantaeb}
\authornotemark[1] 
\orcid{0000-0002-4046-3419}
\affiliation{%
  \institution{University of Amsterdam}
  \city{Amsterdam}
  \country{The Netherlands}}
\email{z.abbasiantaeb@uva.nl}

\author{Yifei Yuan}
\orcid{0000-0001-7275-5398}
\affiliation{%
  \institution{University of Copenhagen}
  \city{Copenhagen}
  \country{Denmark}}
\email{yiya@di.ku.dk}

\author{Mohammad Aliannejadi}
\orcid{0000-0002-9447-4172}
\affiliation{%
  \institution{University of Amsterdam}
  \city{Amsterdam}
\country{The Netherlands}}
\email{m.aliannejadi@uva.nl}

\author{Maarten de Rijke}
\orcid{0000-0002-1086-0202}
\affiliation{%
  \institution{
  University of Amsterdam}
  \city{Amsterdam}
  \country{The Netherlands}}%
\email{m.derijke@uva.nl}

\renewcommand{\shortauthors}{Siro and Abbasiantaeb et al.}

\begin{document}

\title[The Role of Images in Clarifying Questions] {Do Images Clarify? A Study on the Effect of Images on Clarifying Questions in Conversational Search}

\begin{abstract}
\Acf{CS} systems increasingly employ clarifying questions to refine user queries and improve the search experience. Previous studies have demonstrated the usefulness of text-based clarifying questions in enhancing both retrieval performance and user experience. While images have been shown to improve retrieval performance in various contexts, their impact on user performance, when incorporated into clarifying questions, remains largely unexplored. 
We conduct a user study with 73 participants to investigate the role of images in \ac{CS}, specifically examining their effects on two search-related tasks: (i) answering clarifying questions, and (ii) query reformulation. 
We compare the effect of multimodal and text-only clarifying questions in both tasks within a \ac{CS} context from various perspectives. Our findings reveal that while participants showed a strong preference for multimodal questions when answering clarifying questions, preferences were more balanced in the query reformulation task. The impact of images varied with both task type and user expertise: in answering clarifying questions, images helped maintain engagement across different expertise levels, while in query reformulation, they led to more precise queries and improved retrieval performance. Interestingly, for clarifying question answers, text-only setups demonstrated better user performance as they provided more comprehensive textual information in the absence of images. These results provide valuable insights for designing effective multimodal \ac{CS} systems, highlighting that the benefits of visual augmentation are task-dependent and should be strategically implemented based on the specific search context and user characteristics.

\end{abstract}



\maketitle

\acresetall

\section{Introduction}

Understanding a user's query and intent is one of the main challenges in \ac{IR}. Users' queries are often very short and can be interpreted in various ways. Search result diversification~\cite{clarke-2008-novelty} is a traditional solution to this problem; it aims to present results that cover various aspects or interpretations of the same query. In \ac{CS}, however, due to limited bandwidth, search result diversification is not an effective approach. Other interaction modes and mixed-initiative strategies are employed to enhance the system's understanding of user intent~\cite{Radlinski2017Theoretical}, such as preference elicitation, asking clarifying questions, feedback, and query reformulation.
Asking clarifying questions is typically used to find out the intent behind the user's query~\cite{AliannejadiSigir19,siro-agentcq}. 

Much work has been done on understanding the impact of clarifying questions both on system and user performance~\cite{Krasakis2020AnalysingTE,Zamani2020AnalyzingAL,Zou2023Users}. Studies on text-only clarifying questions show that asking just one clarifying question can lead to considerable improvements in retrieval performance~\cite{AliannejadiSigir19,Wang2021Controlling}. 
Clarifying questions are an important element of the user experience in both conversational and ad-hoc retrieval, as determined through controlled user studies~\cite{Kiesel2018TowardVQ,Zou2023Users} and large-scale log analyses~\cite{Zamani2020AnalyzingAL}. User studies on text-only questions~\cite{Zou2023Users,Kiesel2018TowardVQ} lead to multifold findings, where the usefulness of clarifying questions highly depends on their quality, as well as the user's prior knowledge about the task, greatly impacting user satisfaction. 

\begin{figure}
    \centering
    \includegraphics[width=0.8\linewidth]{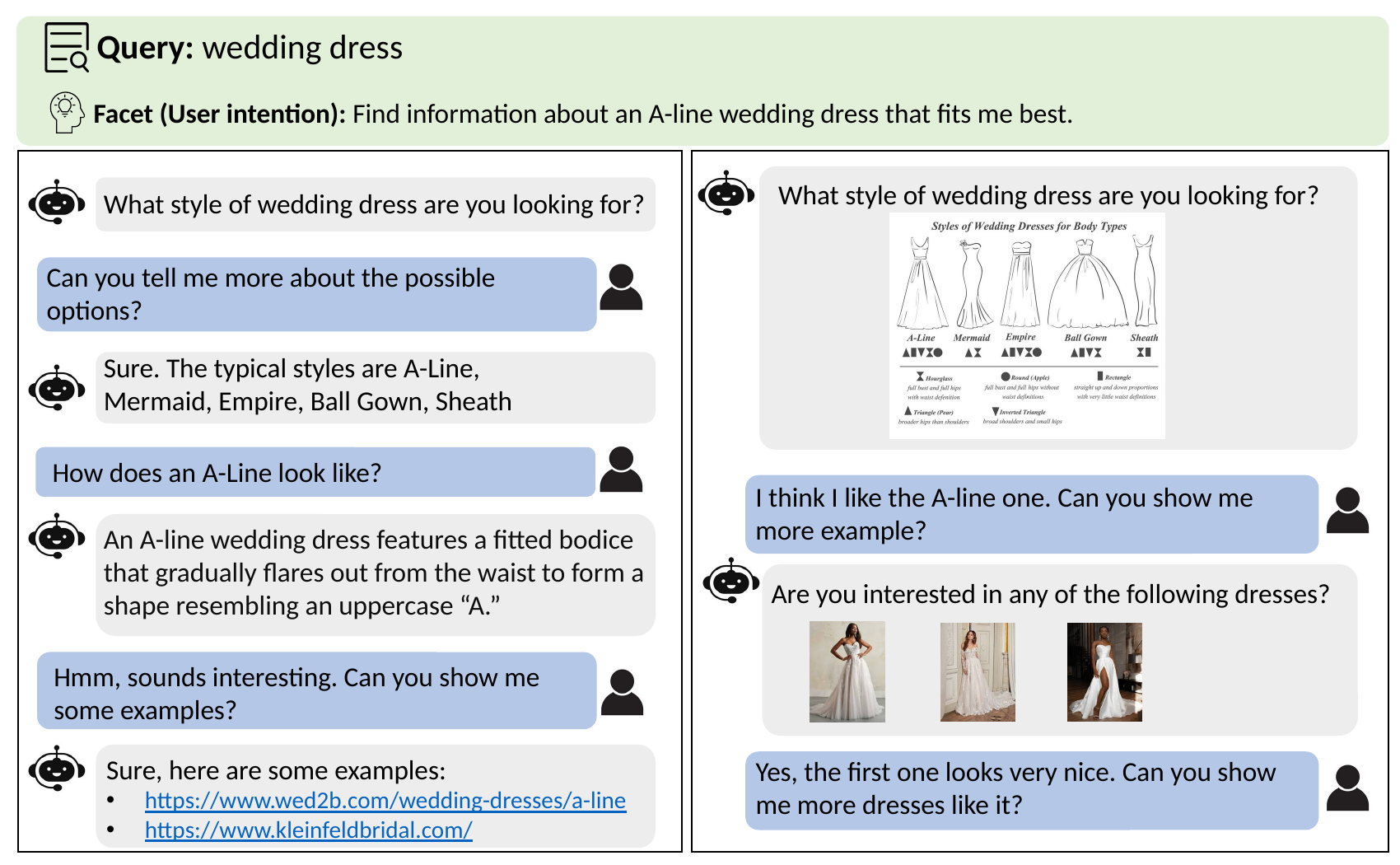}
    \caption{Two example conversations with text-only clarifying questions, representing the existing systems (on the left), as well as multimodal clarifying questions, representing our proposed system setup (on the right).}
    \label{fig:example}
\end{figure}

While clarifying questions enhance \ac{CS} effectiveness, text-only approaches face limitations in handling queries with visual attributes.  Users often struggle to interpret and respond to questions about visual concepts, spatial relationships, or physical attributes through text alone. For instance, describing the specific style of dress you are seeking (~Figure~\ref{fig:example}) or explaining the symptoms of a skin condition can be challenging without visual reference points -- common elements in domains such as medical diagnosis, product search, and architectural design~\citep{DBLP:conf/wsdm/DiSPB14,li2020picture,HOUTS2006173}.

Multimodal search research has demonstrated that visual elements can significantly enhance the search process by reducing cognitive load, providing immediate context, and enabling faster recognition of relevant information compared to text-only descriptions~\citep{fichter2008image,DBLP:conf/sigir/DaganGN21}. Building on these insights, recent work has explored multimodal clarifying questions in \ac{CS}, where systems augment textual questions with relevant images to provide additional context and facilitate user understanding~\cite{Yuan2024Asking}. As illustrated in Figure~\ref{fig:example}, such visual enhancements can provide crucial contextual information while potentially influencing users' perception of system understanding.

While previous research has examined the system-side benefits of multimodal clarifying questions, understanding user interaction behavior and experience remains crucial for developing effective multimodal \ac{CS} systems. We address this gap by investigating how the visual enhancement of clarifying questions affects both user experience and their performance.
Specifically, we examine how users perceive and use image-enhanced clarifying questions across two fundamental search tasks: answering clarifying questions and query reformulation. Through a within-subject controlled study, we present participants with search scenarios comprising an initial query (e.g., ``wedding dress'') and its corresponding information need (e.g., ``Find information about the wedding dress that fits me best''). Participants interact with clarifying questions (e.g., ``What style of wedding dress are you looking for?'') under both with-mage and without-image conditions, providing responses aligned with the given information need (e.g., ``I want to know more about A-line styled wedding dresses'').
This experimental design allows us to systematically investigate how visual enhancement influences user behavior, satisfaction, and performance across different search tasks and user expertise levels.

Our study addresses the following research questions:
\begin{enumerate}[leftmargin=*,label=\textbf{RQ\arabic*}]
\item How do images influence users' answers to clarifying questions in \ac{CS}? \label{rq1}
\item What effect do images have on query reformulation in \ac{CS}? \label{rq2}
\item When are images useful in \ac{CS}? \label{rq3}
\end{enumerate}

\noindent%
We examine these questions in different types of search tasks and levels of user expertise. Our findings reveal several important patterns in how visual elements influence search interaction. First, the impact of images varies significantly between search tasks: while users strongly prefer multimodal clarifying questions for direct question answering, their preferences are more nuanced during query reformulation. Second, we find that visual elements play a crucial role in bridging expertise gaps: images help maintain engagement across different knowledge levels in answering clarifying questions, whereas text-only questions show declining effectiveness as user expertise increases. Third, our analysis reveals an interesting disconnect between user preferences and their performance: while users generally prefer image-enhanced interactions, their performance (measured by the retrieval effectiveness of their answers) varies by task type. In query reformulation, images help users generate more precise queries, but in question answering, text-only responses often lead to better retrieval outcomes as users provide more comprehensive textual information.

Our study and findings provide useful insights into the design and use of images in multimodal \ac{CS} systems. Overall, users find the images useful and helpful in the clarification process; however, their performance reveals mixed results where images are more useful for certain types of search tasks, and clarifying questions. 
This suggests that when deciding to add an image to a clarifying question, the system should take into account the nature of the user's query, as well as the question type, as they greatly impact the usefulness of images.

\section{Related Work}

\subsection{User intent clarification}
Asking clarifying questions enables IR systems to collect users’ explicit feedback, making it an effective interaction mode for various applications,
such as product search~\cite{Zou2019Learning}, voice queries~\cite{Kiesel2018TowardVQ}, question answering~\cite{Braslavski2017WhatDY,Xu2019AskingCQ,Rao2018LearningTA}, and information-seeking systems~\cite{AliannejadiSigir19,Hashemi2020GuidedTL,Zamani2020GeneratingCQ}.
In mixed-initiative IR systems, where either system or the user can take the initiative~\cite{Hearst1999TrendsC,Allen1999MixedinitiativeI}, text-only user intent clarification has been studied extensively~\cite{Zamani2020AnalyzingAL,Sekulic2021UserEP,Owoicho2023ExploitingSU}. 
It is known that, in mixed-initiative systems, clarifying questions have the potential to improve search quality and user experience~\cite{Vakulenko2020AnAO,Meng2021InitiativeAwareSL,Aliannejadi2021AnalysingMI}. 

The research highly depends on user-system interaction data. \citet{AliannejadiSigir19} present a set of clarifying questions and their answers on TREC Web track~\cite{Collins2014Web} queries, which has later been extended to include more topics and clarification need labels, as part of the ConvAI3 competition~\cite{aliannejadi-etal-2021-building}. 
\citet{Zamani2020MIMICSAL} propose a template-based question generation framework and release a dataset based on the questions generated by their algorithm after being deployed at Bing, called MIMICS.
\citet{Sekulic2021User} propose a model trained on the MIMICS dataset, to predict user engagement for a given clarifying question, and use it as a proxy of quality.
Knowing when to ask clarifying questions is an important problem~\cite{Aliannejadi2020ConvAI3GC}, as it can negatively affect the system performance~\cite{Krasakis2020AnalysingTE} and user experience~\cite{Zou2023Users}. 
\citet{Aliannejadi2021AnalysingMI} simulate user-system interactions in a conversational information-seeking scenario and study different strategies where they find that depending on the strategy and user's interaction preference, different numbers of clarifying questions can lead to a better cost-gain trade-off. 
\citet{Wang2021Controlling} argue that in various cases, abstaining from asking clarifying questions leads to better retrieval performance (on top of less user effort) and propose a reinforcement-learning-based algorithm to ask a question (or not) based on the expected information gain.

While these studies have established the importance of clarifying questions in \ac{CS}, they predominantly focus on text-based interactions. The potential of visual elements in the clarification process remains largely unexplored, despite the recognized benefits of multimodal interaction in other search contexts~\citep{turk-2014-multimodal}. Our work extends this body of research by examining how visual elements influence both the clarification process and user responses, contributing new insights into multimodal \ac{CS}.

\subsection{Multimodal IR}
Multimodal IR aims at improving the user experience and system performance by incorporating multimodal information in the interface and retrieval process~\cite{Wang2024What}, applied in various IR scenarios such as query reformulation~\cite{Yuan2022McQueenAB}, question answering~\cite{Chang2021WebQAMA,Talmor2021MultiModalQACQ}, and cross-modal retrieval~\cite{Radford2021LearningTV}. 
In mixed-initiative systems, \citet{MurrugarraLlerena2018ImageRW} propose an image retrieval system that dynamically decides whether the system or user's initiative would be more beneficial to the system's performance. \citet{Ma2021MixedModalityII} combine mixed-initiative and mixed-modal interactions to improve user experience while interacting with conversational recommender systems. 
\citet{Yuan2024Asking} propose integrating images in the clarification process and releases a multimodal query clarification dataset on the task, based on ClariQ. They show that incorporating images in the clarification process leads to significant retrieval improvements.

Our work differs from these works as they focus on system performance or other IR applications. Our work, on the other hand, focuses on the effect of images on user experience and performance in mixed-initiative \ac{CS} systems.

\subsection{User aspects}
A lot of research has focused on system and utility aspects of asking text-only and multimodal clarifying questions. As a relatively recent means of user-system interaction, various user-related aspects of asking clarifying questions are yet to be studied. \citet{Kiesel2018TowardVQ} conduct one of the first user studies on the impact of asking clarifying questions on user performance and experience in systems dealing with voice queries. They find that users find such an interaction helpful even in cases where it does not provide a helpful interaction experience. 
\citet{Azzopardi2022Towards} propose a theoretical framework based on an economics model of IR, accounting for various interaction modes involving clarifying questions.
In an attempt to discover the effect of clarifying questions of different qualities on the user experience, \citet{Zou2023Users} run a large-scale controlled study where they simulate the clarification system deployed by Bing~\cite{Zamani2020GeneratingCQ} and present the participants with questions of different qualities. 
They find that low-quality questions can lead to worse user experience and performance. 
Presenting multiple questions in the same search session can reduce the risk. 
In a follow-up study, they examine the effect of clarifying questions in multi-question sessions. 
Although it is less risky to ask multiple clarifying questions in the same session, they find that users start to lose their trust in the system if the system starts the session with low quality questions~\cite{Zou2023Users}.

While inspiring our study, these studies do not focus on image-enhanced clarifying questions. Also, in most cases, the information-seeking task mimics a web search scenario, whereas we focus on multimodal conversational information-seeking.

\section{Study Design}
To investigate the role images play during search clarification, we conducted a controlled user study. We examined the impact of images on \ac{CS} interactions, focusing on two key tasks: answering clarifying questions and query reformulation. We selected these tasks as they represent the primary actions users take during \ac{CS} clarification -- either directly answering system questions or modifying their queries based on the interaction~\cite{Aliannejadi2021AnalysingMI}. The tasks represent two distinct yet interconnected aspects of the search process. When users respond to clarifying questions, they must interpret the system's request and provide relevant information, a process that may be altered by the presence of visual cues. Similarly, query reformulation represents users' evolved understanding of their information need, potentially influenced by the clarifying question and its accompanying image(s).
Based on the prior work on \ac{CS} systems~\citep{AliannejadiSigir19} and mixed-initiative search~\citep{DBLP:journals/pacmhci/AvulaCA22}, we propose the following hypotheses:

\begin{enumerate}[label=\textbf{H\arabic*},leftmargin=*]
\item Multimodal clarifying questions will lead to higher user engagement and satisfaction, compared to text-only questions. \label{h1}
\begin{description}[leftmargin=*]
\item[\textit{Rationale:}] Visual elements could provide context that complements textual information in clarifying questions. A multimodal approach would enhance user understanding and engagement by offering multiple channels for processing information. The complementary nature of images and text would could help maintain efficient task completion times. The integration of visual and textual elements aligns with cognitive theories suggesting that multiple representational formats can enhance information processing and comprehension~\citep{mayer2003nine}.
\end{description}

\item Users' background knowledge on a search topic will influence their perception and dependence on images in clarifying questions. \label{h2}
\begin{description}[leftmargin=*]
\item[\textit{Rationale:}] 
Based on research showing that domain knowledge significantly affects how users process and integrate information during search tasks~\citep{DBLP:conf/wsdm/WhiteDT09}, we hypothesize that experts and novices may differ in their ability to extract and utilize visual information effectively during the clarification process.

\end{description}

\item The utility of images in clarifying questions will vary based on query type, with higher perceived usefulness for queries with inherent visual attributes (e.g., descriptive or visual information needs) than abstract or conceptual queries. \label{h3}
\begin{description}[leftmargin=*]
\item[\textit{Rationale:}] Visual content naturally excels at conveying physical and spatial characteristics that text alone struggles to describe efficiently~\citep{turk-2014-multimodal,bobek-2016-creating}. When answering clarifying questions about visual attributes (like product appearances or spatial arrangements), users can reference images directly rather than interpreting textual descriptions~\citep{DBLP:conf/sigir/QuYCTZQ18}. Similarly, during query reformulation, images provide concrete visual anchors that help users articulate visual concepts more precisely in their refined queries~\citep{DBLP:conf/cikm/HuangE09}.
\end{description}
\end{enumerate}

\subsection{Topic selection and pre-study analysis}
To investigate the role of images in clarifying questions, we need to carefully curate a set of search tasks where images can have a meaningful impact. Although the ClariQ dataset~\citep{aliannejadi-etal-2021-building} provides a foundation of \ac{CS} topics, it has not been designed with visual elements in mind. 
MELON~\citep{Yuan2024Asking}, an extension of ClariQ incorporating images, has inconsistencies in image quality and relevance due to its crowdsourced image collection process (as revealed by our initial inspection). Previous studies show that crowdworkers may not consistently select optimal images for search tasks~\citep{NowakR10-reliable-annotations}. To control for these variables, we curate our own image collection and conduct a systematic pre-study to identify suitable topics for this study.

In our pre-study task, we sampled 100 topics from ClariQ and employed two appen\footnote{\href{Appen}{https://www.appen.com/}} assessors to judge the potential benefit of image augmentation for clarifying questions. We asked the assessors to provide detailed justifications as why the topics would benefit from visual elements (or not). 
Analysis of workers' justifications revealed three key characteristics that made topics amenable to visual augmentation:

\begin{enumerate}[leftmargin=*]
    \item Physical structures or objects (e.g., ``hip roof construction,'' ``solar water fountains'');
    \item Medical conditions with visual symptoms (e.g., ``carpal tunnel syndrome''); and
    \item Natural elements requiring visual identification (e.g., ``norway spruce characteristics'').
\end{enumerate}

\noindent%
Based on these insights, we selected 24 topics~(Table~\ref{table:require_image_no_image}) and formulated six additional ones, making it a total of 30 topics for the study. We modified the information needs to align with identified visual enhancement opportunities while maintaining the original search context from ClariQ. For example, ``hip roof'' is extended to address visual aspects of roof structure and design elements. Images are sourced from Google image search by querying the topic, aiming to complement the clarifying questions.

\begin{table}[t]
\centering
\caption{Topics and facets selected from ClariQ for the study with the reference to ClariQ facet ids in brackets. Note: \small{topics without a reference are those that we reformulated ourselves.
}}
\begin{tabular}{ll}
\toprule
\textbf{Topics} & \textbf{Topics} \\ \midrule
T1: dangers of asbestos (F0075)                 & T16: volvo (F0459)                           \\ 
T2: norway spruce (F0736)                       & T17: land surveyor (F0455)                   \\ 
T3: home theater systems (F0468)                & T18: american military university (F0146)    \\ 
T4: grilling (F0243)                            & T19: ct jobs (F0889)                         \\ 
T5: dinosaurs (F0162)                           & T20: cass county missouri (F0032)            \\ 
T6: kids earth day activities (F0491)           & T21: electoral college 2008 results (F0507)  \\ 
T7: teddy bears (F0607)                         & T22: rick warren (F0497)                     \\ 
T8: hip roof (F0616)                            & T23: angular cheilitis (F0206)               \\ 
T9: solar water fountains (F0493)               & T24: barrett's esophagus (F0600)             \\ 
T10: carotid cavernous fistula treatment (F0716) & T25: moths (F0921)                          \\ 
T11: ham radio (F0543)                          & T26: patron saint of mental illness (F0609) \\ 
T12: carpal tunnel syndrome (F0477)             & T27: altitude sickness (F0757)              \\ 
T13: cloud types &  T28: car dashboard symbol \\
T14: Bike repair &  T29: office chair \\
T15: coffee table &  T30: pipe fittings \\
\bottomrule
\end{tabular}
\label{table:require_image_no_image}
\end{table}

\subsection{Study setup}
We employed a within-subjects design. Both studies used the same set of 30 topics, with each topic presented in two setups: with- and without-image. This design allowed for a direct comparison of how the presence or absence of an image affected participants' responses to the same questions. 

\subsubsection{Task structure}
For each task, participants were presented with:
\begin{enumerate}[leftmargin=*]
\item \textbf{Information need:} A detailed description of the user's context and search goal, derived from the ClariQ facets.
\item \textbf{Initial query:} The first search query a user entered, based on their information need.
\item \textbf{Clarifying question:} A question from the system, designed to better understand the user's intent or narrow down the search focus.
\item \textbf{Image:} An image added to clarifying question, intended to provide additional context or information.  
\end{enumerate}

\subsubsection{Tasks} We explain the tasks below:

\header{Task 1: Answering clarifying questions}
We asked the participants to imagine themselves as the actual users seeking information and to answer the clarifying question as if they had the given information need.

\header{Task 2: Query reformulation}
We asked the participants to act as users with the information need who had submitted their initial query. Their task was to reformulate their initial query after being exposed to the clarifying questions.

\subsubsection{Questionnaire design}
Following established practices in interactive IR evaluation~\citep{DBLP:journals/ftir/Kelly09}, we designed three questionnaires to measure  aspects of the user experience with multimodal clarifying questions:

\header{Pre-task demographics questionnaire}
Before beginning the tasks, participants completed a questionnaire capturing:
\begin{enumerate}[leftmargin=*]
\item Demographic information (age, gender, and education level),
\item Experience with conversational systems (5-point scale: no experience to very experienced), and
\item Experience with the task.
\end{enumerate}
The demographic questions were selected based on factors known to influence search behavior~\citep{DBLP:conf/wsdm/WhiteDT09} and to ensure our sample represented diverse user characteristics.

\header{Post-clarification questionnaire}
After each clarifying question, participants completed a brief questionnaire designed to capture immediate feedback. The choice of questions was motivated by aspects of \ac{CS} interaction~\citep{aliannejadi-etal-2021-building}:
\begin{enumerate}[leftmargin=*]
\item Information sources relied upon when answering the clarifying question (e.g., image, initial query, clarifying question, and personal knowledge).%
\item Background knowledge level on the topic (3-point scale: expert, familiar, new topic).
\item Clarity of the provided clarifying question (5-point scale: very unclear to very clear).%
\item Usefulness of the provided clarifying question (5-point scale: not useful at all to very useful).%
\item Overall satisfaction (5-point scale: very poor to excellent).%
\end{enumerate}

\header{Exit questionnaire}
We designed the exit questionnaire
to capture overall patterns across the interaction. Questions covered:
\begin{enumerate}[leftmargin=*]
\item Overall experience with multimodal clarifications (5-point scale: significantly difficult to significantly easy).
\item Overall experience with text-only clarifications (5-point scale: significantly difficult to significantly easy).
\item Impact of images on clarification efficiency (5-point scale: greatly slowed to greatly sped up).
\item Frequency of cases where images provided helpful context (5-point scale: never to always).
\item Preference for setup (multimodal, text-only, and no preference).
\item Justification for their preference (open-ended).
\item Situations where images were most helpful in understanding clarifying questions (e.g., physical objects and processes\footnote{These categories were derived from our pre-study analysis of topic characteristics and aligned with the types of queries where visual elements were predicted to be most beneficial.}).
\end{enumerate}

\subsection{Procedure}
The study was conducted in two separate experiments (i.e., answering clarifying questions and query reformulation), each with its own set of participants. Both experiments included the following steps before starting the study of main search tasks as shown in Figure~\ref{fig:example}: 
(i) pre-task test and consent to the collection of their demographic information; 
(ii) filling in the demographic information; 
and
(iii) reading instructions and examples on completing the task. 

\begin{figure}
    \centering
    \includegraphics[width=0.8\textwidth]{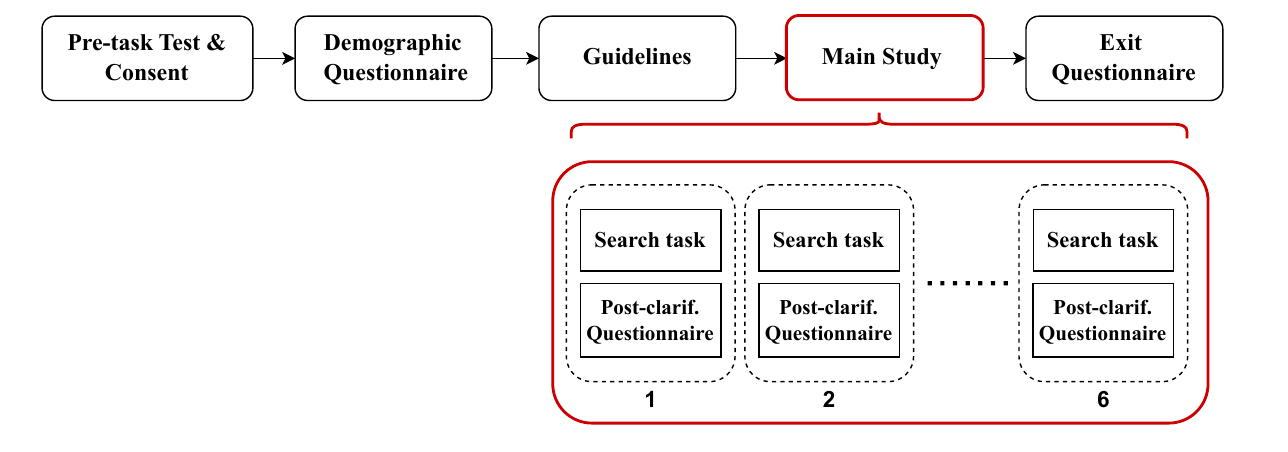}
    \caption{Our user study procedure. ``Main Study'' refers to either query reformulation or answering the clarifying question tasks.}
    \label{fig:enter-label}
\end{figure}

\subsubsection{Search task distribution}
In total, we had 60 search tasks for each experiment. We randomly grouped them into 10 batches where each batch included six search topics. We ensured that each participant encountered an equal number of multimodal and text-only clarifying questions and that each clarifying question was evaluated in both setups across the participant pool. Each batch was assigned to at least 3 users. To mitigate learning effects, we limited each participant's involvement to a maximum of two batches for a single experiment. While participants were exposed to both with- and without-image clarifying questions, we structured the batches to ensure that no participant encountered the same clarifying question in both setups within a single batch. The users were given the tasks one by one. After successful completion and submission of each task, they were enabled to move to the next task.
The entire procedure for each experiment was designed to be completed within approximately 10--15 minutes. The detailed procedure for each experiment is as follows:

\header{Task 1: Answering clarifying questions}
Each participant engaged with six search topics. The order of the questions and the presence/absence of images were randomized and counterbalanced to mitigate learning effects. After each question, participants completed the post-clarification questionnaire. Upon responding to all six questions, participants answered the exit questionnaire.

\header{Task 2: Query reformulation and exploration}
A different set of participants responded to six search topics, each involving query reformulation. As in Task 1, the order of questions and the presence/absence of images was randomized and counterbalanced. Participants completed the post-clarification questionnaire after each question and the exit questionnaire after completing all six questions.

\subsubsection{Quality assurance}
To ensure the quality of our user study we implemented the following:
\begin{enumerate}[leftmargin=*]
    \item We started the study with a pre-task test including two attention check questions. Participants were allowed to only do this test once. In case of failure, participants were not allowed to continue with the study. In case of successful completion of the pre-task test, participants were redirected to the main task.
    \item We provided a comprehensive guideline for the participants to complete the tasks. We included detailed descriptions and examples.
    \item After completing the study by participants, we manually checked the responses provided by them and discarded the low-quality annotations. We discarded low-quality responses from six participants in Task 1 and seven in Task 2.
    \item We ran three pilot studies to improve our guidelines and questions and in each pilot, we used four participants. 
    \item We maintained the integrity of the experiment by recruiting separate groups of participants for each task. Participants in Task 1 were not eligible to participate in Task 2, and vice versa. This approach prevented potential bias and cross-contamination between the two tasks.
    
\end{enumerate}

\subsection{Data collection}
We collected the following data:
\begin{itemize}[leftmargin=*]
\item Text responses (answers to clarifying questions, reformulated queries);
\item Self-reported information reliance;
\item Ratings on various aspects from the questionnaires (pre- and post-clarification, and exit questionnaires); and
\item Open-ended responses from the exit questionnaire (justification on image preference).
\end{itemize}
Overall, we collected 360 data points; 180 for each task. Per task, we collected 90 samples per condition (with- vs.\ without-image).
All collected data is stored in a local password-protected computer to ensure data privacy.

\subsection{Participants}
We recruited participants through Prolific.\footnote{\url{https://www.prolific.com/}} To ensure high-quality responses, we applied strict filters, requiring participants to have a 95\% or higher approval rate and to have completed more than 3000 tasks on Prolific. Eligible participants were 18 years or older, had English as their native language, and were regular users of search engines and digital assistants. 

Task 1, which focused on answering clarifying questions, involved 36 unique participants ($N = 36$), each participant completed one batch. The age distribution was: 38--47 years ($n = 10$), 28--37 and 48--57 years ($n = 7$ each), 68+ years ($n = 6$), 58--67 years ($n = 5$), and 18--27 years ($n = 1$). Male participants ($n = 25$) outnumbered female ($n = 11$) participants. Most participants held bachelor's degrees ($n = 25$), followed by master's degrees ($n = 6$), PhDs ($n=2$), and other ($n = 3$). The majority ($n = 27$) had no prior experience in answering clarifying questions, while some ($n = 9$) reported previous experience. Participants reported having moderate knowledge levels on the topics involved in the task ($2.59 / 3$).

Task 2 involved 37 unique participants ($N = 37$), each completing one batch. The age distribution was: 38--47 years ($n = 10$), 48--57 and 58--67 years  ($n = 8$ each), 28--37 years ($n = 6$), 18--27 years ($n = 4$), 68+ years ($n = 1$). There were 21 males, 13 females and 3 preferred not to say. Bachelor's degree holders formed the largest group ($n = 26$), followed by master's degree holders ($n = 6$), while high school graduates, PhD holders, and those with other qualifications each comprised $n = 2$ participants. Regarding task experience in reformulating clarifying questions, $n = 24$ participants reported no prior experience, while $n = 13$ indicated previous experience. For query reformulation, participants reported having an average knowledge level of 2.6 on the search topics.

\subsection{Data analysis}
Our study employed a mixed-methods approach to analyze the collected data, combining quantitative statistical analyses with qualitative content analysis. 

\subsubsection{Quantitative analysis}
The quantitative analysis examined responses from both post-clarification and exit questionnaires using multiple statistical approaches. To account for the hierarchical nature of our data, where responses are nested within participants, we employed linear mixed-effects models with setup (with- vs.\ without-image) as a fixed effect and participant as a random effect. This approach allowed us to control for individual participant variations while examining the effect of image presence.
To complement the mixed-effect analysis, we conducted independent t-tests comparing responses between with- and without-image setups, with Bonferroni correction ($\alpha = 0.05/6 = 0.0083$) to control for multiple comparisons. Effect sizes were calculated using Cohen's $d$. We conducted one-way ANOVA tests with post-hoc Tukey's HSD to examine the influence of participants' background knowledge levels (novice, familiar, and expert) on their interactions.
All statistical analyses were performed with a significance level of $\alpha = 0.05$, with appropriate corrections for multiple comparisons where applicable.

\subsubsection{Qualitative analysis}
The qualitative analysis examined three types of open-ended responses: answers to clarifying questions, reformulated queries, and justifications for setup preferences. Two independent coders analyzed these responses using thematic analysis. The coders first independently identified recurring patterns in how participants used images across different response types. They then developed a coding scheme through discussion and iteration, focusing on patterns that emerged in both with- and without-image conditions. The final coding scheme was applied to all responses.

\section{Results}

In this section, we present findings from our user study examining both tasks: answering clarifying questions (Task 1) and query reformulation (Task 2). We analyze results across two conditions (with- and without-image).

\subsection{Descriptive statistics}
\begin{figure}[t]
\centering
\begin{subfigure}{.35\textwidth}
\centering
\includegraphics[width=.95\linewidth]{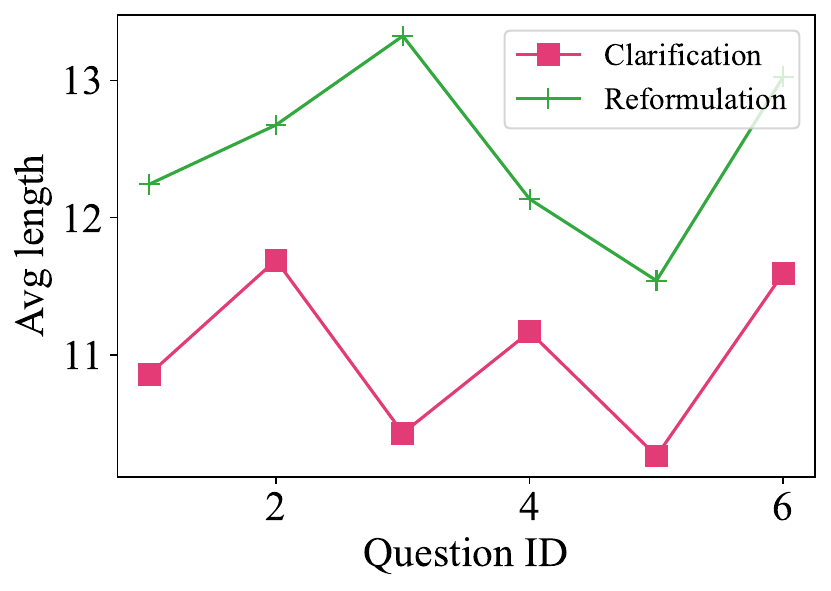}
\caption{Average answer length  per question}
\label{fig:length}
\end{subfigure}%
\begin{subfigure}{.35\textwidth}
\centering
\includegraphics[width=.95\linewidth]{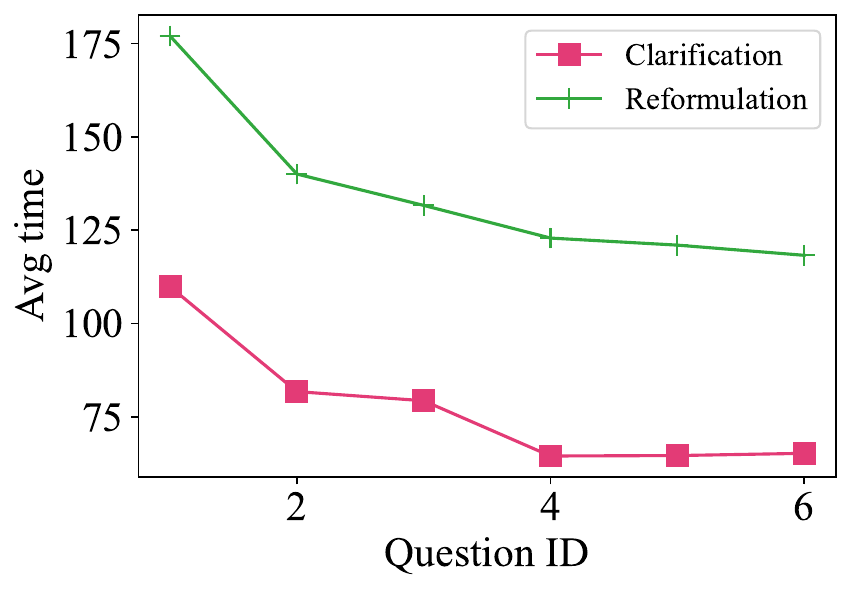}
\caption{Average completion time per question}
\label{fig:time}
\end{subfigure}

\caption{Average time taken by participants to complete each question (a) and the average length of the clarifying question and reformulated queries~(b).}
\label{fig:time-length}
\end{figure}

Participants spend on average 77.52 seconds answering each clarifying question in Task 1; Task 2 requires substantially more time with an average of 135.15 seconds per query reformulation. This difference highlights the increased cognitive demand for query reformulation compared to direct question answering. Analysis of completion times reveals a consistent decrease as participants progressed through the questions (Figure~\ref{fig:time}). To distinguish between potential fatigue and learning effects, we examine the relationship between completion time and response length. While time decreases across questions, the average length of both answers and reformulated queries showed consistent variation (Figure~\ref{fig:length}), suggesting that faster completion times result from task familiarity rather than decreased engagement.

The exit questionnaire completion times also reflect the differential cognitive load between tasks: Task 1 takes 84.50 seconds (median: 69.00 seconds) and Task 2 takes 92.14 seconds (median: 74.00 seconds), with a mean difference = 7.64 seconds, and median difference = 5.00 seconds. 
This difference suggests that the nature of the preceding task (answering clarifying questions vs. reformulating queries)
influences participants’ response times and potentially their depth of reflection. We hypothesize that the longer
completion times in Task 2 may be due to the complexity of the task, thus leading to more extensive cognitive processing, not only during the task itself but also during subsequent reflection.

\subsection{Task 1: Answering clarifying questions}
Table~\ref{table:mean_ratings1} presents the results, combining mean rating, mixed-effects analysis, and independent t-test results.

\header{Question clarity} 
Our analysis reveals that clarifying questions are generally perceived as clear across both setups, with ratings predominantly in the 4--5 range (Figure \ref{fig:clarity_distribution1}). Interestingly, questions without images are rated marginally clearer (3.93 vs. 3.85), with our analysis showing a small negative effect of images ($\beta = -0.079$, d = -0.078), where questions without images receive slightly more maximum clarity ratings (5/5).  This raises an interesting possibility that images might occasionally introduce complexity rather than clarity to the question-answering process. The high ICC for clarity (0.420) suggests that participants were moderately consistent in their clarity evaluations across questions.

\header{Question usefulness}
Participants find the clarifying questions useful overall, with ratings skewed towards the higher end of the scale (Figure \ref{fig:usefulness_distribution1}). Questions with images are rated marginally more useful (3.78 vs. 3.73), showing a small positive effect ($\beta = 0.056$, d = 0.049). The substantial consistency in individual ratings (ICC = 0.399) indicates that participants maintain stable opinions about question utility across different scenarios, regardless of the setup

\header{Overall satisfaction} Participants report high satisfaction levels across both setups (mean = 3.68 on a 5-point scale). The presence of images has minimal impact on satisfaction ratings (3.66 with images vs. 3.69 without), with our analysis confirming this negligible difference ($\beta = -0.032$, d = -0.037). The high consistency in individual ratings (ICC = 0.399) indicates that participants maintain stable satisfaction levels across different questions, suggesting that image presence does not substantially alter their overall experience with the clarification process.

\header{User answers}
On average,  participants provide relatively concise answers regardless of the experimental setup, with longer answers when images are included (11.30 words, SD = 6.50) than without images (10.53 words, SD = 5.75). 
With-image questions take slightly longer to complete (79.00 seconds vs. 76.05 seconds), suggesting that while images may prompt more detailed responses, they also require additional processing time.  
The moderate ICC (0.355) for completion time indicates that while individual differences exist in response speed, they are not as pronounced as in other measures.

\subsection{Task 2: Query reformulation}
\begin{figure}[t]
\centering
\begin{subfigure}{.3\textwidth}
\centering
\includegraphics[width=.95\linewidth]{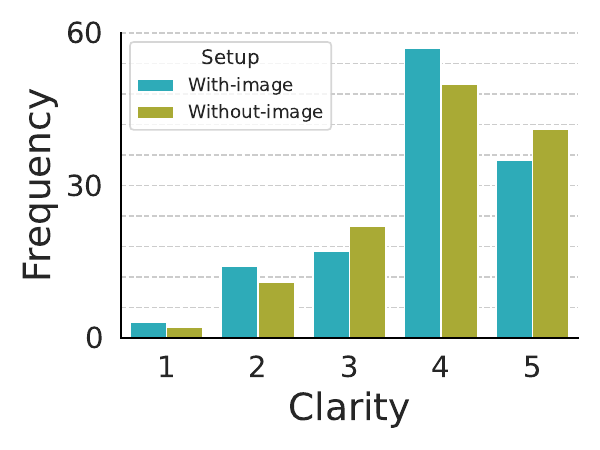}
\caption{Clarity }
\label{fig:clarity_distribution1}
\end{subfigure}
\begin{subfigure}{.3\textwidth}
\centering
\includegraphics[width=.95\linewidth]{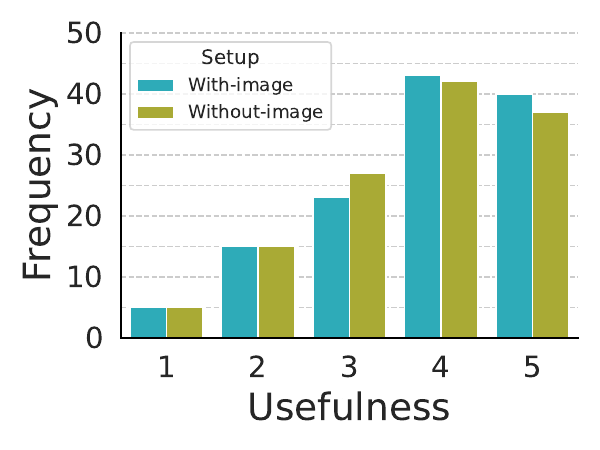}
\caption{Usefulness }
\label{fig:usefulness_distribution1}
\end{subfigure}%
\begin{subfigure}{.3\textwidth}
\centering
\includegraphics[width=.95\linewidth]{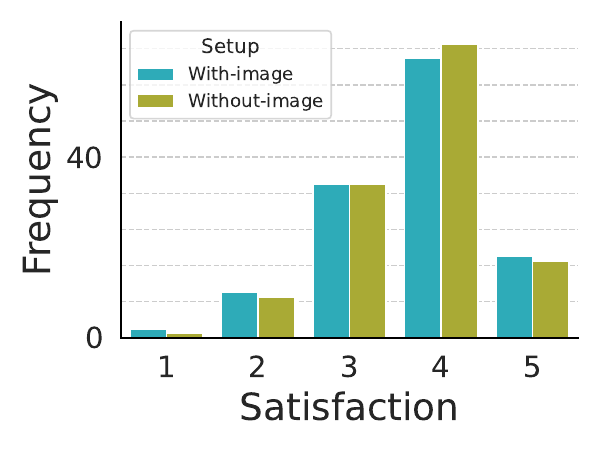}
\caption{Satisfaction level}
\label{fig:satisfaction_experience1}
\end{subfigure}
\begin{subfigure}{.3\textwidth}
\centering
\includegraphics[width=.95\linewidth]{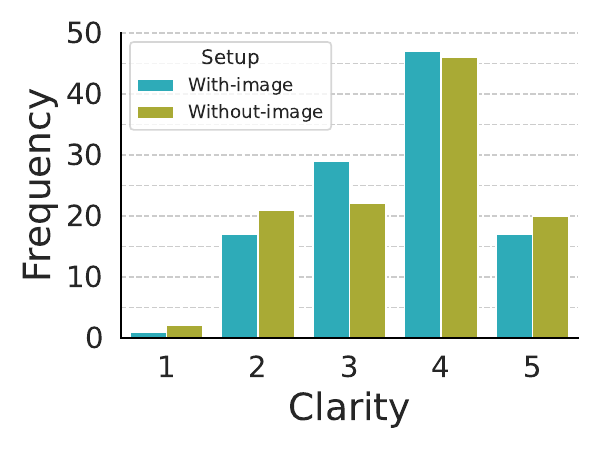}
\caption{Clarity }
\label{fig:clarity_distribution2}
\end{subfigure}
\begin{subfigure}{.3\textwidth}
\centering
\includegraphics[width=.95\linewidth]{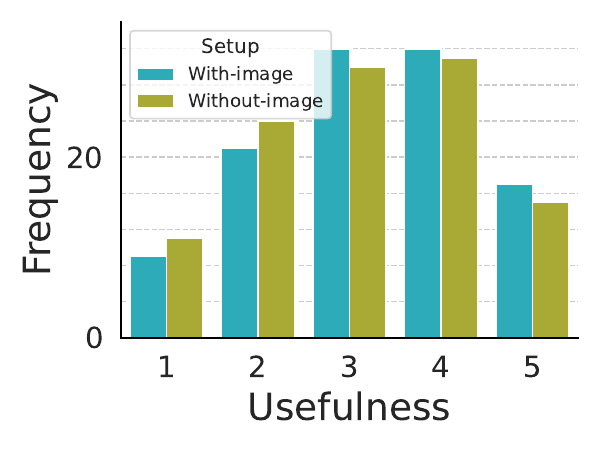}
\caption{Usefulness }
\label{fig:usefulness_distribution2}
\end{subfigure}%
\begin{subfigure}{.3\textwidth}
\centering
\includegraphics[width=.95\linewidth]{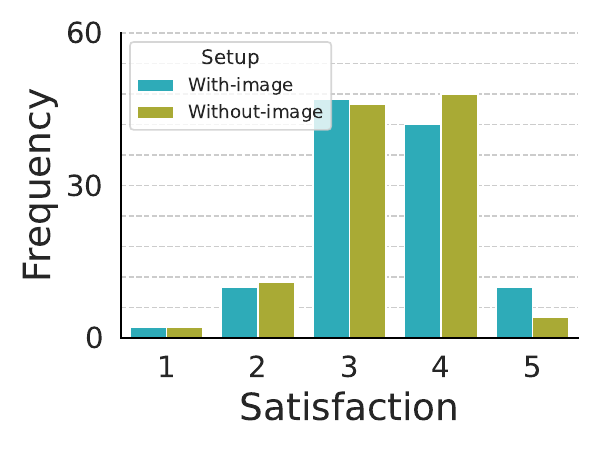}
\caption{Satisfaction level}
\label{fig:satisfaction_experience2}
\end{subfigure}
\caption{Rating distributions of main task aspects as rated by participants in Task 1 -- row 1 and Task 2 -- row 2.}
\label{fig:main-task1}
\end{figure}

Task 2, focusing on query reformulation, shows distinctive patterns in participant ratings for clarity and usefulness, as shown in Figure \ref{fig:main-task1}. 

\header{Question clarity}
While question clarity ratings are positive, Table~\ref{table:mean_ratings2} indicates a minimal difference between setups (3.56 vs.\ 3.55) ($\beta = 0.009$, $d = 0.009$). The lower ICC for clarity (0.286) compared to Task 1 suggests more variability in how participants evaluate clarity in the reformulation tasks.

\header{Usefulness}
Usefulness ratings are moderate with Figure~\ref{fig:usefulness_distribution2}  showing a shift towards lower ratings compared to Task 1. While multimodal questions were rated slightly more useful (3.24 vs.\ 3.14) ($\beta = 0.108$, $d = 0.091$), the higher ICC (0.351) indicates more consistent individual preferences.

\header{Overall satisfaction}
Satisfaction levels remain moderate to high, with a small advantage for multimodal conditions (3.43 vs. 3.37, $\beta = 0.063$, $d = 0.077$). The high consistency in individual ratings (ICC = 0.357) indicates that participants maintain stable satisfaction levels across different reformulation tasks.

\header{Reformulated queries}
For query length, while original queries are quite short (mean = 3.13 words), reformulated queries are significantly longer (mean = 12.46 words). When comparing the two setups, we observe a subtle difference in reformulated query lengths: queries are slightly shorter when images are included (12.29 words vs. 12.63 words). This suggests that images may lead to more concise query reformulations, possibly by helping users focus their information needs more precisely. 

\header{Completion time}
The markedly higher completion time (M = 135.63s, SD = 61.70 with images) compared to Task 1 reveals the more cognitively demanding nature of query reformulation. The high ICC for completion time (0.503) suggests that individual differences in reformulation strategies are more stable and pronounced than in clarifying question answering.

\header{Summary}
Our findings do not support \textbf{H1}, that visually-enhanced clarifying questions lead to higher user satisfaction and engagement. While images show some positive effects on usefulness ratings and minor benefits for satisfaction, these differences are not substantial. The impact of images varies by task type: in Task 1, they lead to longer answers but slightly reduced clarity, while in Task 2, they result in more concise reformulations with minimal impact on clarity. This suggests that the value of visual enhancement may be more nuanced and task-dependent than initially hypothesized.

\begin{table*}[t]
\caption{Analysis results for Task 1: Clarifying Question Answering. {\small Note: Mean rating shows M (SD). $\beta$ represents the fixed effect of condition (image presence); SE = Standard Error of the coefficient estimate.} {\small ICC = Intraclass Correlation Coefficient; d = Cohen's d effect size.}}
\label{table:mean_ratings1}
\begin{tabular}{lcccccccc}
\toprule
& \multicolumn{2}{c}{Mean rating} & \multicolumn{2}{c}{Mixed-Effects} & \multicolumn{3}{c}{Independent t-test} & \\
\cmidrule(lr){2-3} \cmidrule(lr){4-5} \cmidrule(lr){6-8}
Measure & With Image & Without Image & $\beta$ (SE) & ICC & t(180) & p & d & \\
\midrule
Topic Knowledge & 2.60 (0.51) & 2.58 (0.54) & 0.024 (0.061) & 0.160 & 0.360 & 0.719 & 0.045 \\
CQ-Usefulness & 3.78 (1.14) & 3.72 (1.13) & 0.056 (0.124) & 0.243 & 0.389 & 0.698 & 0.049 \\
CQ-Clarity & 3.85 (1.03) & 3.93 (1.00) & \llap{-}0.079 (0.098) & 0.420 & \llap{-}0.622 & 0.535 & \llap{-}0.078 \\
Satisfaction & 3.67 (0.88) & 3.70 (0.82) & \llap{-}0.032 (0.083) & 0.399 & \llap{-}0.297 & 0.767 & \llap{-}0.037 \\
Answer Length & 11.30 (6.50)\phantom{0} & 10.53 (5.75)\phantom{0} & 0.770 (0.596) & 0.407 & 0.996 & 0.320 & 0.125 \\
Completion Time & 79.00 (54.72) & 76.05 (44.92) & 2.952 (5.078) & 0.355 & 0.468 & 0.640 & 0.059 \\
\bottomrule
\end{tabular}
\end{table*}

\begin{table*}[t]
\caption{Analysis Results for Task 2: Query Reformulation. {\small Note: Mean rating shows M (SD). $\beta$ represents the fixed effect of condition (image presence); SE = Standard Error of the coefficient estimate.} {\small ICC = Intraclass Correlation Coefficient; d = Cohen's d effect size.}}
\label{table:mean_ratings2}
\begin{tabular}{lcccccccc}
\toprule
& \multicolumn{2}{c}{Mean rating} & \multicolumn{2}{c}{Mixed-Effects} & \multicolumn{3}{c}{Independent t-test} & \\
\cmidrule(lr){2-3} \cmidrule(lr){4-5} \cmidrule(lr){6-8}
Measure & With Image & Without Image & $\beta$ (SE) & ICC & t(180) & p & d & \\
\midrule
Topic Knowledge & 2.64 (0.54) & 2.56 (0.53) & 0.081 (0.065) & 0.177 & 1.129 & 0.260 & 0.152 \\
CQ-Usefulness & 3.24 (1.17) & 3.14 (1.20) & 0.108 (0.128) & 0.351 & 0.681 & 0.496 & 0.091 \\
CQ-Clarity & 3.56 (0.96) & 3.55 (1.05) & 0.009 (0.114) & 0.286 & 0.067 & 0.947 & 0.009 \\
Satisfaction & 3.43 (0.85) & 3.37 (0.79) & 0.063 (0.088) & 0.357 & 0.575 & 0.566 & 0.077 \\
Refor. Query l & 12.29 (5.26)\phantom{0} & 12.63 (5.37)\phantom{0} & \llap{-}0.342 (0.546) & 0.417 & \llap{-}0.480 & 0.632 & \llap{-}0.064 \\
Completion Time & 135.63 (61.70)\phantom{0} & 134.68 (72.97)\phantom{0} & 0.955 (6.393) & 0.503 & 0.105 & 0.916 & 0.014 \\
\bottomrule
\end{tabular}
\end{table*}

\subsection{Information sources relied on when answering clarifying questions and reformulating queries}
\begin{figure}[t]
\centering
\begin{subfigure}{.5\textwidth}
\centering
\includegraphics[width=.95\linewidth]{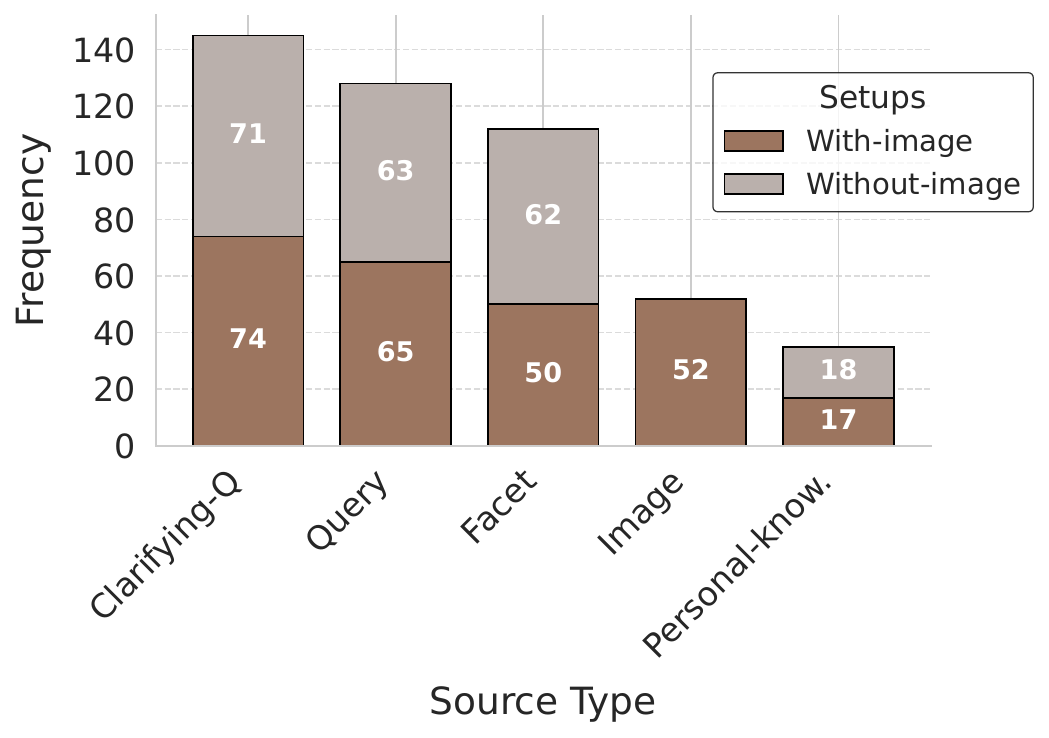}
\caption{Task 1}
\label{fig:reliedino1}
\end{subfigure}%
\begin{subfigure}{.5\textwidth}
\centering
\includegraphics[width=.95\linewidth]{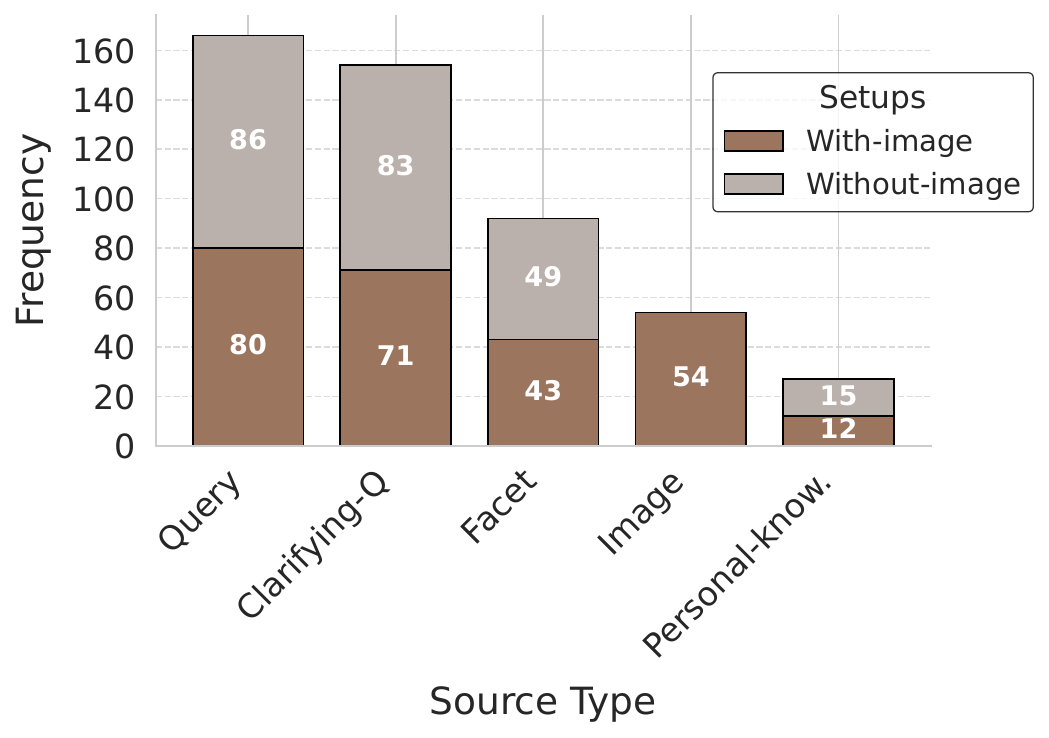}
\caption{Task 2 }
\label{fig:reliedinfo2}
\end{subfigure}
\caption{Information sources relied on by participants when a) answering clarifying questions and b) reformulating queries.}
\label{fig:main-task-reliedinfo}
\end{figure}

Figure~\ref{fig:main-task-reliedinfo} presents participants' utilization of information sources across two search tasks: answering clarifying questions (Task 1) and query reformulation (Task 2), comparing with-image and without-image setups.

\header{Task-based patterns}
The primary information source shifts between tasks, with clarifying questions dominating in Task 1 (145 total references) and initial queries in Task 2 (166 references). This aligns with task requirements: question answering naturally emphasizes the clarifying question, while reformulation centers on modifying the original query. Task 2 generally exhibits higher frequencies of information source usage than Task 1, particularly for clarifying questions and initial queries. On average each participant relies on 3.2 information sources in Task 2 compared to 1.8 in Task 1. This difference likely reflects the more complex cognitive process involved in reformulating a query, which requires synthesizing information from multiple sources to generate a new, improved query. 

\header{Impact of visual enhancement}
The presence of images influences information source utilization patterns. In Task 1, visual cues enable a more balanced use of clarifying questions (74) and queries (65), while its absence leads to increased facet use (62 vs.\ 50). Task 2 shows stronger compensation patterns in the without-image condition, with higher reliance on both queries (86 vs.\ 80) and clarifying questions (83 vs.\ 71).

Across both tasks, personal knowledge remains consistently the least referenced source, with minimal variation between conditions, suggesting that participants primarily relied on provided information rather than prior knowledge regardless of visual support.

\header{Summary} These patterns demonstrate that the presence of images alters how users approach information gathering in \ac{CS}. While images serve as an additional information source, their impact extends beyond direct usage -- they appear to streamline the information integration process and reduce reliance on textual sources. This suggests that visual elements not only provide direct information but also help users more efficiently process and combine information from multiple sources during search tasks.

\subsection{Exit questionnaire aspects}

\begin{figure}[t]
\centering
\begin{subfigure}{.45\textwidth}
\centering
\includegraphics[width=.95\linewidth]{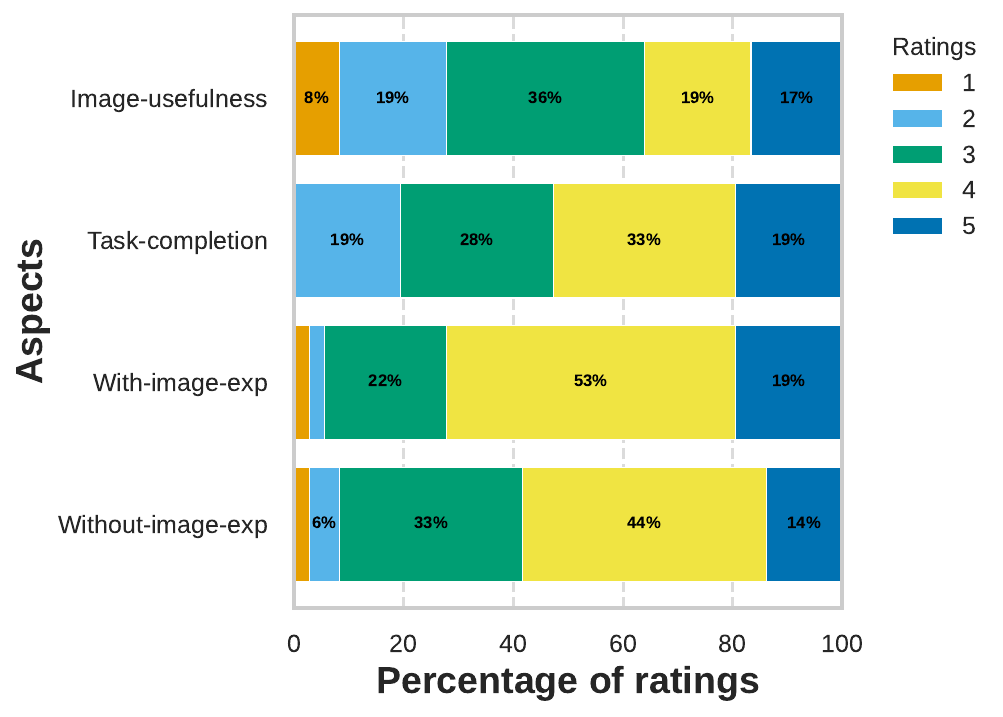}
\caption{Task 1}
\label{fig:task1-exit}
\end{subfigure}%
\begin{subfigure}{.45\textwidth}
\centering
\includegraphics[width=.95\linewidth]{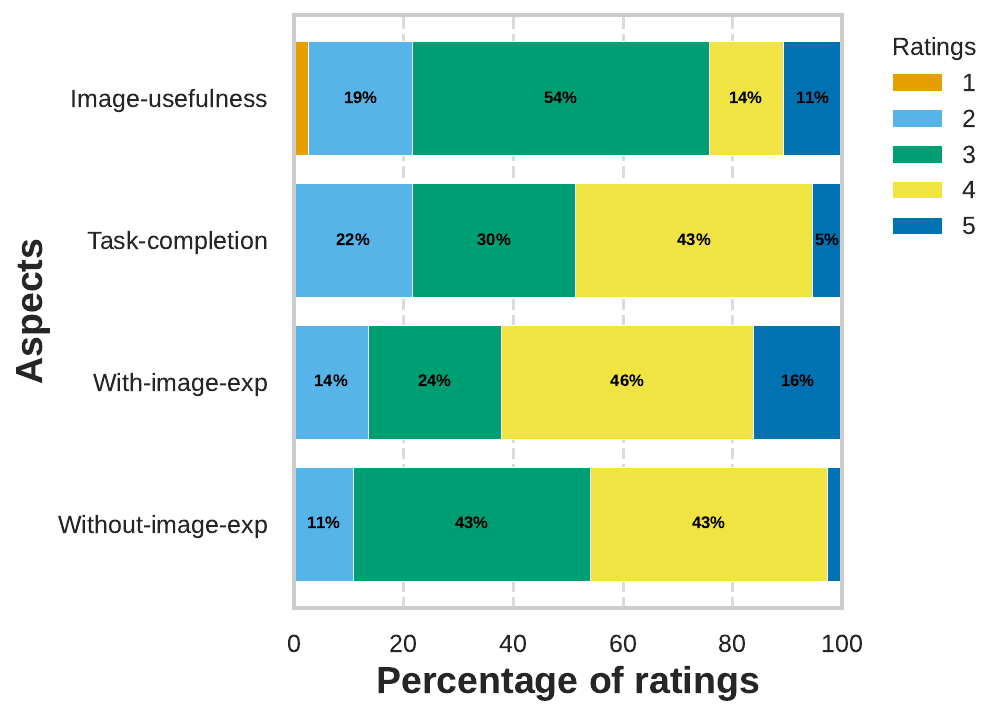}
\caption{Task 2}
\label{fig:task2-exit}
\end{subfigure}
\caption{Rating distributions of final task aspects as rated by participants in Task 1.}
\label{fig:final-task-stacked}
\end{figure}

The exit questionnaire survey results, depicted in Figure~\ref{fig:final-task-stacked}, reveal participants' perceptions and preferences regarding multimodal clarifying questions across both tasks.

\begin{figure}[t]
\centering
\begin{subfigure}{.4\textwidth}
\centering
\includegraphics[width=.95\linewidth]{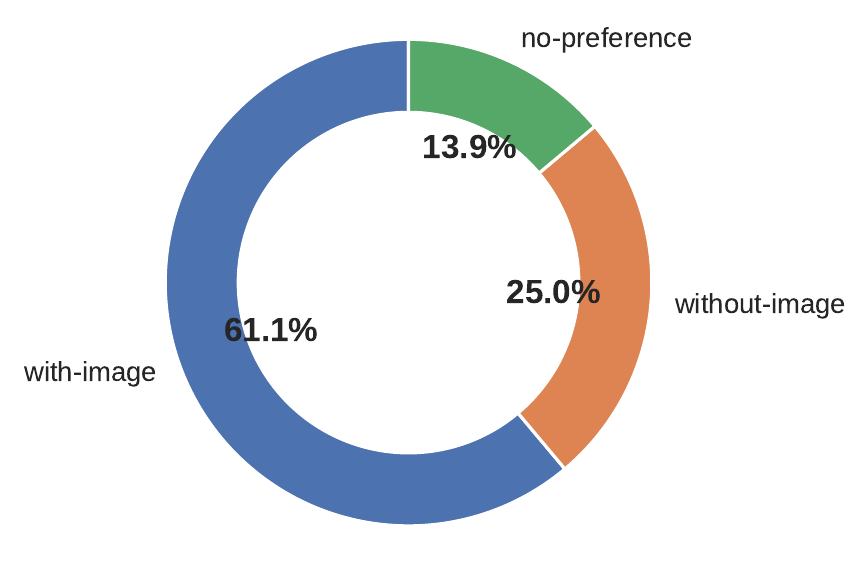}
\caption{Task 1}
\label{fig:pref-task1}
\end{subfigure}%
\begin{subfigure}{.4\textwidth}
\centering
\includegraphics[width=.95\linewidth]{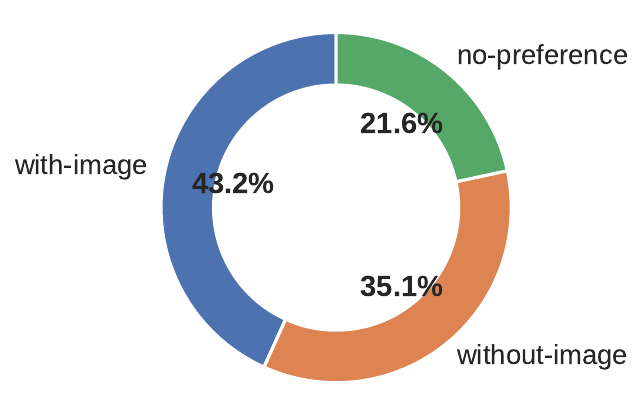}
\caption{Task 2}
\label{fig:pref-task2}
\end{subfigure}
\caption{Preference of setup.}
\label{fig:final-taskpref}
\end{figure}

\header{Task 1} In Task 1, images significantly enhance the question-answering process. 36\% of participants find images ``Sometimes'' helpful, with 33\% reporting improved task understanding and completion speed. Users strongly prefer multimodal setups, with 61.1\% favoring this setup (Figure~\ref{fig:pref-task1}), reflected in the high percentage rating on their experience with image setup. The ease of use is evident, with 72\% rating multimodal setups as easy or significantly easy. These results demonstrate that visual aids provide immediate context and facilitate comprehension when answering clarifying questions. The straightforward nature of this task allows users to directly leverage visual information, explaining the consistent, moderately beneficial impact of images.

\header{Task 2} Task 2, images prove more consistently useful, with 68\% rating them as ``Sometimes'' or ``Often'' helpful~(Figure~\ref{fig:task2-exit}). 43\% of users report improved task understanding and completion speed with images. However, preferences are more evenly distributed: 43.2\% prefer multimodal setups, 35.1\% favor text-only setups, and 21.6\% have no preference. This balanced distribution aligns with the polarized experience ratings for multimodal setups in Task 2. These findings indicate that while users recognize the potential value of images in providing rich, multi-dimensional context that can spark new ideas or highlight previously unconsidered aspects of a topic, the actual preference for using images in this task is more varied. This variation stems from the complex nature of query reformulation, which requires synthesizing information and generating new search terms. While, images may provide rich, multi-dimensional context that can spark new ideas, the same richness can be distracting for some users or query types.

\header{Summary}
These findings provide partial support for \textbf{H1} (visually-enhanced clarifying questions lead to higher user satisfaction and engagement). While images enhance satisfaction in the question-answering task, their impact on query reformulation is more variable. This suggests that the effectiveness of visual enhancement depends on the specific task context: straightforward for direct question answering, but more complex and individual-dependent for query reformulation. The higher reported usefulness but lower preference for images in Task 2 indicates that visual aids can provide valuable context while potentially introducing cognitive complexity in more demanding tasks.

\subsection{Image usefulness in query clarification} \label{sec:img-usefulness}

\begin{figure}[t]
\centering
\begin{subfigure}{.45\textwidth}
\centering
\includegraphics[width=.95\linewidth]{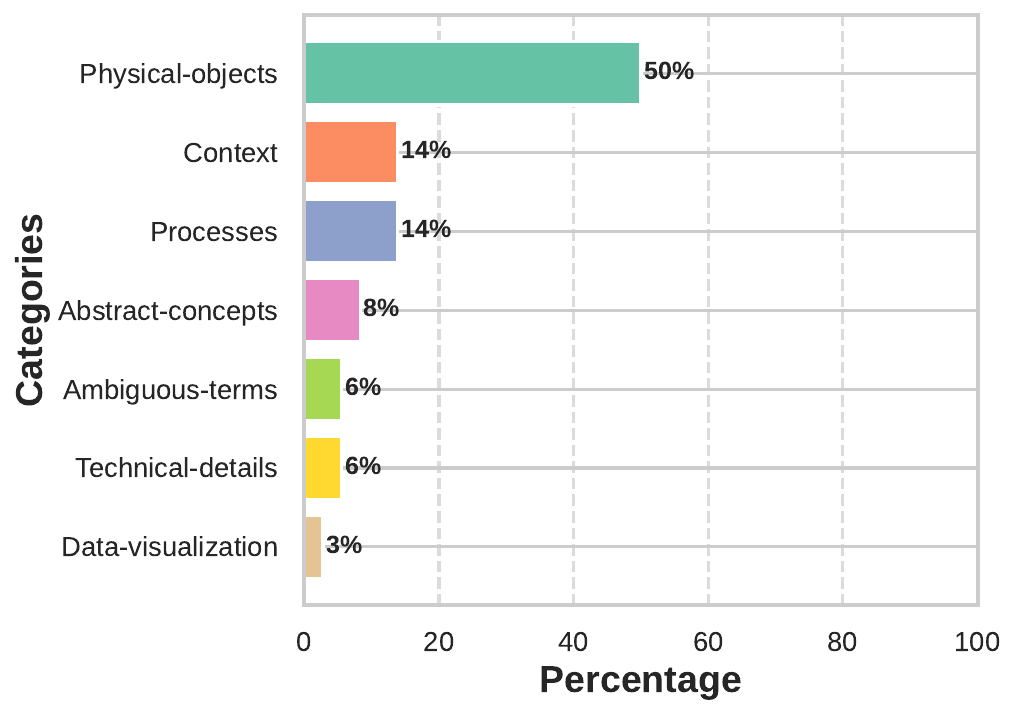}
\caption{Task 1}
\label{fig:setup1_distribution1}
\end{subfigure}%
\begin{subfigure}{.45\textwidth}
\centering
\includegraphics[width=.95\linewidth]{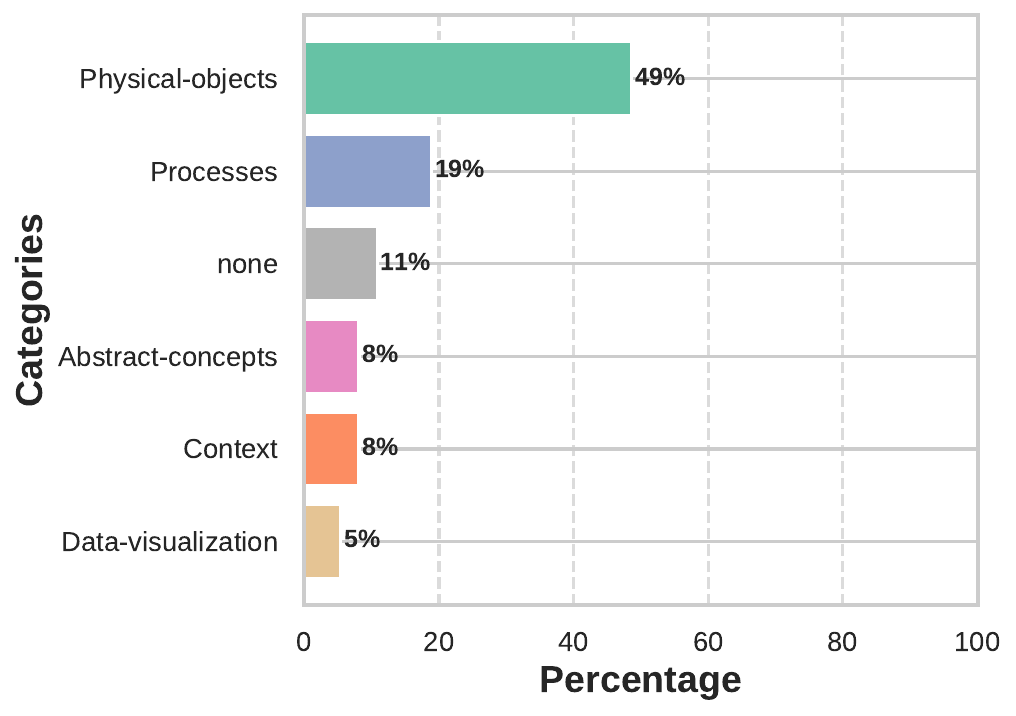}
\caption{Task 2}
\label{fig:setup2_distribution1}
\end{subfigure}
\caption{Participants self-reported situations where images were useful.}
\label{fig:final-task-situations}
\end{figure}

Figure \ref{fig:final-task-situations} reveals the distribution of situations where participants find images to be useful; Table \ref{tab:image_utility} has examples.
In Task 1, physical object scenarios dominate at 50\%. This high percentage indicates that when users seek information about tangible items, such as dinosaurs (T5), images are likely to be highly beneficial. Context-related and process-related situations each account for 14\%, demonstrating equal importance in scenarios like clarifying hip roof construction (T8) or explaining grilling techniques (T4). Abstract concepts represent 8\% of situations, while technical details and data visualization account for 6\% and 3\%, respectively.

Task 2 maintains the prominence of physical object scenarios at 49\%, showing the consistent value of visual representations for tangible items. It reveals a significant increase in process-related situations to 19\%. This increase reflects the greater utility of visual demonstrations when users refine queries about specific processes, such as detailed grilling methods (T4). Context-related scenarios decrease to 8\%, indicating less need for broad visual overviews during query reformulation. Abstract concepts remain steady at 8\%; data visualization shows a slight increase to 5\%.

To better understand when images are perceived as useful or not, we analyze participants' justifications for their setup preferences through thematic analysis. Our analysis reveals four primary useful aspects and four not-useful aspects of image utility in search interactions (Table \ref{tab:thematic}).

\header{Beneficial aspects}
Images prove most valuable for contextual support (28.4\% of participants), particularly when users need additional context to understand and reformulate their queries. As one participant notes, ``The images helped me to have some more context to rephrase the question.'' Cognitive facilitation is the second most prominent benefit (24.3\%), with participants highlighting how images simplify information processing and reduce cognitive load, especially for complex topics.
Creative stimulation (21.6\%) is another key benefit, with images inspiring alternative query formulations and new perspectives: ``They made me think of alternative angles for the questions that I might not have thought about otherwise.'' Enhanced engagement and focus (16.2\%) round out the beneficial aspects, with images helping maintain task attention and improve understanding.

\header{Limiting aspects}
Our analysis also reveals limitations. Cognitive overload from irrelevant visuals (13.5\%) is the most frequently cited problem, particularly when images introduce unnecessary complexity. Some participants (10.8\%) find images irrelevant to their task goals, noting misalignment between visual content and search objectives. Individual preferences play a role, with 9.5\% expressing a clear preference for text-based interaction. Finally, 8.1\% view images as redundant, adding no value beyond the textual information.

\header{Summary} Our findings provide strong support for \textbf{H3}: image utility varies systematically with query type. The clear dominance of physical objects and process-related scenarios in perceived image usefulness, coupled with the lower utility reported for abstract concepts, confirms our hypothesis. Our thematic analysis further reinforces this pattern, showing that positive perceptions of images (contextual enhancement, cognitive efficiency) are predominantly associated with queries having inherent visual attributes, while negative perceptions (cognitive overload, irrelevance) are more common with abstract or conceptual queries. 

\begin{table}[h]
\centering
\small
\caption{Examples of topics and clarifying questions for different image utility situations.}
\label{tab:image_utility}
\begin{tabular}{p{2.5cm}p{4.75cm}p{6.75cm}}
\toprule
\textbf{Situation} & \textbf{Topic (T)} & \textbf{Clarifying Question (CQ)} \\
\midrule
Physical objects & T5: Dinosaurs & CQ1: Are you interested in the types or their history? \\
\midrule
Processes & T4: Grilling & CQ2: What techniques or tips for grilling vegetables would you like to know more about? \\
\midrule
Context & T8: Hip roof & CQ3: What specific information do you need regarding the construction or specifications of a hip roof? \\
\midrule
Abstract concepts & T10: Carotid cavernous fistula treatment & CQ4: What particular treatment options are you most curious about? \\
\midrule
Technical details & T11: Ham radio & CQ5: Are you looking for information on the different designs of ham radio antennas and their specific applications? \\
\midrule
Data visualization & T12: Carpal tunnel syndrome & CQ6: Which types of exercises are you considering—those for immediate relief or for long-term prevention? \\
\bottomrule
\end{tabular}

\end{table}

\begin{table*}[t]
\caption{Thematic analysis of Participants setup preference justifications to understand when images are useful and not}
\label{tab:thematic}
\small
\begin{tabular}{p{2.8cm}p{5cm}p{5cm}@{}c}
\toprule
\textbf{Category} & \textbf{Characteristics} & \textbf{Example quote} & \textbf{Freq. (\%)} \\
\midrule
\multicolumn{4}{l}{\textit{Useful aspects}} \\
\midrule
Contextual enhancement & Provides additional context; Enhances query understanding; Clarifies search intent & ``Gave me additional information to enable me to re-formulate my question'' & 28.4 \\
\midrule
Cognitive efficiency & Reduces mental effort; Simplifies information processing; Enables faster comprehension & ``A picture very often helps clarify things without a lot of written, and sometimes superfluous information'' & 24.3 \\
\midrule
Creative stimulation & Inspires new query angles; Suggests alternative formulations; Promotes exploration & ``They made me think of alternative angles for the questions that I might not have thought about otherwise'' & 21.6 \\
\midrule
Increased engagement and focus & Increases task engagement; Enhances attention; Improves task focus & ``I thought the images added a good initial visual prompt this helped in determining context'' & 16.2 \\
\midrule
\multicolumn{4}{l}{\textit{Not Useful Aspects}} \\
\midrule
Cognitive overload  & Introduces unnecessary complexity; Creates processing burden; Complicates task & ``They overloaded me with information and I found it more difficult to construct a straightforward question'' & 13.5 \\
\midrule
Irrelevant to user & Misaligns with user expectations; Lacks task relevance; Provides unhelpful information & ``Images were not always relevant to what I needed to know'' & 10.8 \\
\midrule
User preference for text & Individual preference for textual information; Comfort with text-based search & ``I am better with words than images'' & 9.5 \\
\midrule
Redundancy  & Adds no additional value; Duplicates textual information & "I didn't see the images as providing any more information that I required to reformulate the query'' & 8.1 \\
\bottomrule
\multicolumn{4}{p{0.96\textwidth}}{\small Note: percentages represent the proportion of participants who mentioned each aspect in their responses. Individual responses often contained multiple aspects, hence percentages sum to more than 100\%. For example, a single participant might mention both contextual benefits and cognitive facilitation in their response.}
\end{tabular}
\end{table*}

\subsection{Effect of background knowledge}
\begin{figure}[t]
\centering
\begin{subfigure}{.33\textwidth}
\centering
\includegraphics[width=\linewidth]{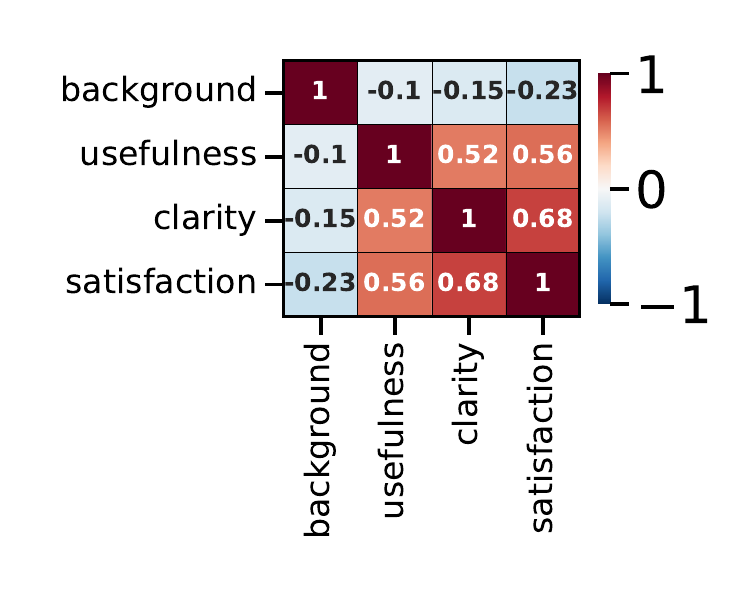}
\caption{All data}
\label{fig:maintask1-corr}
\end{subfigure}%
\hfill
\begin{subfigure}{.33\textwidth}
\centering
\includegraphics[width=\linewidth]{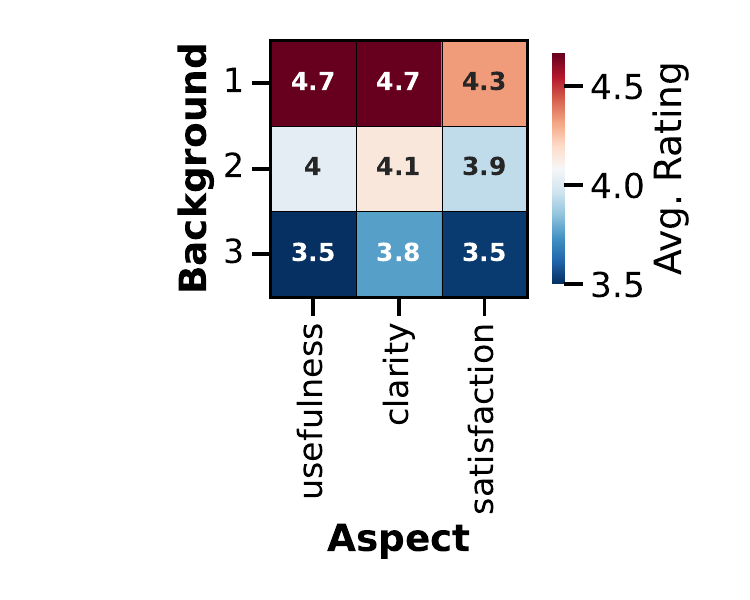}
\caption{Without-image}
\label{fig:maintask1-corr1}
\end{subfigure}%
\hfill
\begin{subfigure}{.33\textwidth}
\centering
\includegraphics[width=\linewidth]{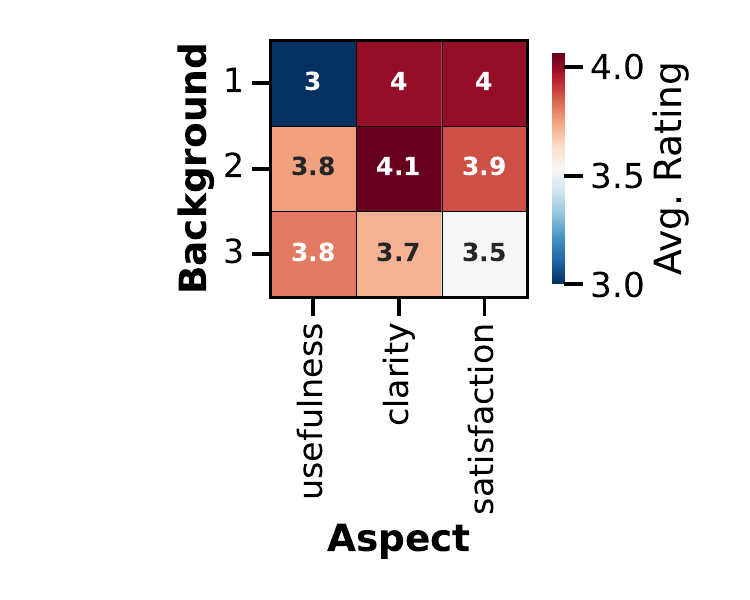}
\caption{With image}
\label{fig:maintask1-corr2}
\end{subfigure}
\begin{subfigure}{.33\textwidth}
\centering
\includegraphics[width=\linewidth]{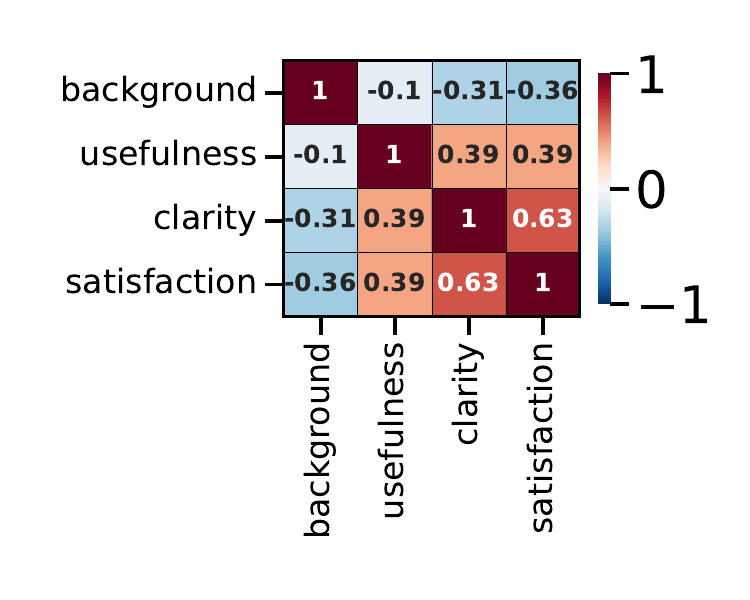}
\caption{All data}
\label{fig:maintask2-corr}
\end{subfigure}%
\hfill
\begin{subfigure}{.33\textwidth}
\centering
\includegraphics[width=\linewidth]{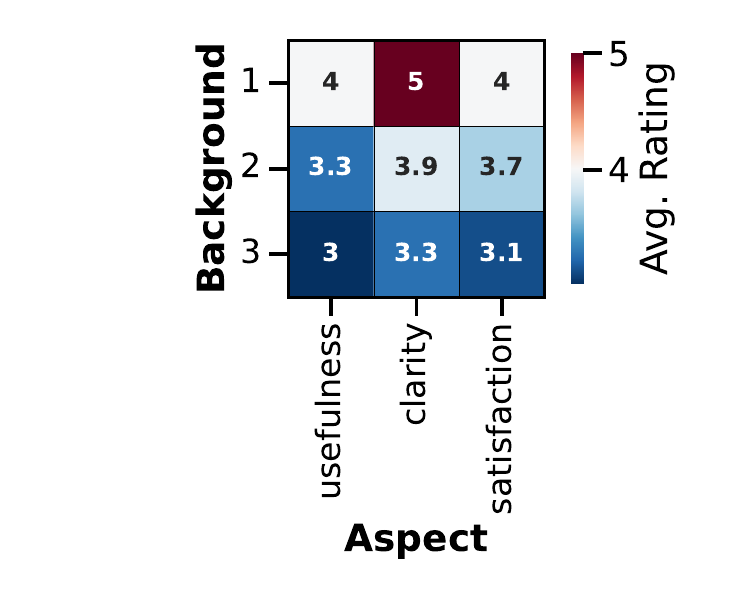}
\caption{Without-image}
\label{fig:maintask2-corr1}
\end{subfigure}%
\hfill
\begin{subfigure}{.33\textwidth}
\centering
\includegraphics[width=\linewidth]{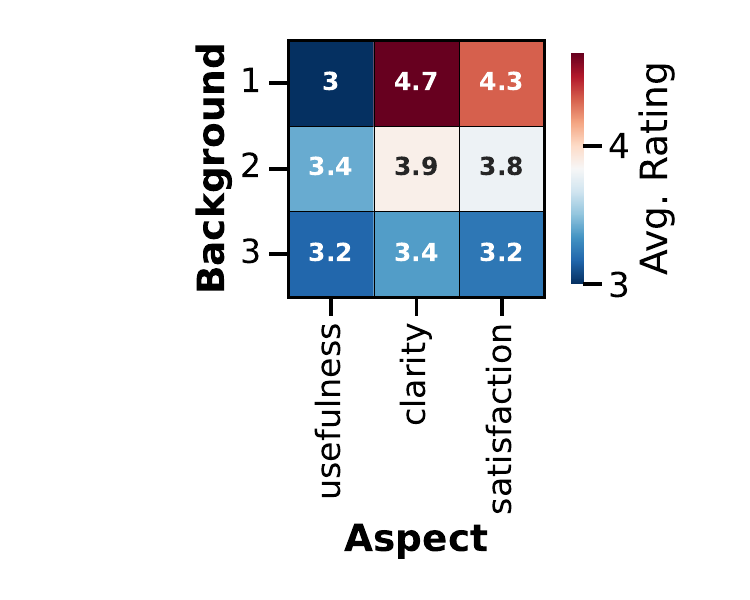}
\caption{With-image}
\label{fig:maintask2-corr2}
\end{subfigure}
\caption{Spearman's $\rho$ correlations for main task questionnaire questions. Top row -- Task 1 and bottom row -- Task 2}
\label{fig:main-task-corr}
\end{figure}

We examine how users' background knowledge influences their perception of clarifying questions with and without images. Figure \ref{fig:main-task-corr} presents Spearman's $\rho$ correlations and mean ratings for usefulness, clarity, and satisfaction across different knowledge levels for both tasks.

In Task 1, while measures show strong positive correlations among themselves (Figure~\ref{fig:maintask1-corr}, $\rho$ > 0.6, p < 0.01), we observe a consistent negative correlation between background knowledge and all other measures. Similarly, Task 2 exhibits even stronger negative correlations between background knowledge and user ratings (Fig. \ref{fig:maintask2-corr}, $\rho$ < -0.4, p < 0.01). Given these observed negative correlations, we conduct further statistical analysis to understand this relationship. For each task, we perform one-way ANOVA tests separately for with and without-image setups, followed by Tukey's HSD for post-hoc comparisons.

\header{Task 1} The without-image setup~(Figure \ref{fig:maintask1-corr1}) reveals significant effects of background knowledge (F(2,97) = 8.45, p < 0.05). Post-hoc comparisons show significant differences between medium and high knowledge levels, with usefulness ratings dropping from 4.7 to 3.5 and satisfaction from 4.3 to 3.5. The with-image setup~(Figure \ref{fig:maintask2-corr1}) shows no significant differences across knowledge levels (F(2,97) = 2.13, p = 0.12), suggesting visual aids help maintain engagement regardless of expertise.

\header{Task 2} For Task 2, without-image setup (Figure \ref{fig:maintask2-corr1}), we find significant knowledge effects on both satisfaction (F(2,97) = 9.23, p < 0.05) and clarity (F(2,97) = 10.11, p < 0.05), with ratings decreasing substantially for higher expertise levels (clarity: 5.0 to 3.3; satisfaction: 4.0 to 3.1). While the with-image setup~(Figure \ref{fig:maintask2-corr2}) maintained significant effects (F(2,97) = 7.84, p < 0.05), the decline was less pronounced (clarity: 4.7 to 3.4; satisfaction: 4.3 to 3.2).

\header{Summary} The consistency of these patterns across tasks, despite their different cognitive demands, strongly supports \textbf{H2}. The findings suggest that background knowledge plays a crucial role in how users perceive clarifying questions, with three key implications:

\begin{itemize}[leftmargin=*]
    \item Expert users generally find text-only clarifying questions less valuable, possibly due to their advanced understanding of the domain;
    \item Images serve as knowledge mediators, providing additional context that remains valuable even at higher expertise levels; and 
    \item The moderating effect of images is particularly important in complex tasks like query reformulation, where the gap between novice and expert perceptions is largest.
\end{itemize}

\noindent%
This suggests that multimodal clarifying questions may be crucial for creating universally effective search experiences, as they provide layered context that can be interpreted differently based on user expertise. The stronger moderating effect in Task 2 also indicates that visual support becomes more valuable as task complexity increases.

\section{Retrieval Effectiveness}
Building on our analysis of user interactions with multimodal clarifying questions, we examine whether visual enhancement impacts retrieval performance. Our evaluation focuses on whether the presence of images in clarifying questions leads to responses and reformulations that improve document retrieval.

We employ BM25 as our retrieval model for several key reasons. First, as a robust, training-free model for text-based ranking, it eliminates potential biases from training data. Second, by focusing on text-based retrieval, we avoid the complexity of aligning text and image features that multimodal models would require, allowing us to directly assess how visual elements influence user-generated text (responses and reformulations). Our experimental setup evaluates three query configurations:
\begin{itemize}[leftmargin=*]
\item Original query alone,
\item Original query augmented with clarifying question-answer pairs, and
\item User-provided reformulated queries
\end{itemize}

\noindent%
For each configuration, we retrieve the top 100 documents per facet and evaluate performance using NDCG@k ($k \in \{1,5,10\}$). Tables \ref{tab:bm25-ret} and \ref{tab:bm25-ret1} present comparative results across with-image and without-image conditions.

\begin{table}[!t]
\centering
\caption{BM25 retrieval and ranking performance with different clarifying questions and answers. Ori. and Refor. denote original and reformulated query, respectively.}
\setlength{\tabcolsep}{3mm}
\begin{tabular}{llccc}
\toprule
& \textbf{Input} & \textbf{NDCG@1} & \textbf{NDCG@5} & \textbf{NDCG@10} \\
\midrule
\multirow{2}{*}{\textbf{Ori.}}
& Query only & 0.059 & 0.067 & 0.074\\
&Query + QA  & 0.142 & 0.130 & 0.148\\
\midrule
\multirow{2}{*}{\textbf{Refor.}}
& Query only & 0.127 & 0.129 & 0.149 \\
&Query + QA & \textbf{0.162} & \textbf{0.136} & \textbf{0.168}\\
\bottomrule
\end{tabular}
\label{tab:bm25-ret}
\end{table}

\begin{table}[!t]
\centering
\caption{BM25 retrieval and ranking performance on with / without image queries with different clarifying questions and answers. Ori. and Refor. denote original and reformulated query, respectively.}
\setlength{\tabcolsep}{3mm}
\begin{tabular}{llccc}
\toprule
& \textbf{Input} & \textbf{NDCG@1} & \textbf{NDCG@5} & \textbf{NDCG@10} \\
\midrule
\multirow{2}{*}{\textbf{Ori.}}
& Query only & 0.059 / 0.059  & 0.067 / 0.067 & 0.074 | 0.074\\
&Query + QA  & 0.079 / 0.185 & 0.120 / 0.152 & 0.138 / 0.172\\
\midrule
\multirow{1}{*}{\textbf{Refor.}}
& Query only & \textbf{0.204} / 0.091 & \textbf{0.155} / 0.120 & \textbf{0.182} / 0.140\\
\bottomrule
\end{tabular}
\label{tab:bm25-ret1}
\end{table}

\subsection{Retrieval performance findings}

Analysis of Tables~\ref{tab:bm25-ret} and~\ref{tab:bm25-ret1} reveals four key patterns in retrieval performance as explained below.

\header{Clarifying questions}
Incorporating clarifying questions significantly enhances retrieval performance compared to using queries alone, consistent with findings from previous studies~\cite{AliannejadiSigir19,Sekulic2021UserEP,Yuan2024Asking}. This improvement demonstrates the value of clarification in enriching search context.

\header{Query reformulation}
Reformulated queries consistently outperform original queries in retrieval effectiveness, aligning with recent research~\cite{Owoicho2023ExploitingSU,Wang2023GenerativeQR}. This suggests that the reformulation process helps users better articulate their information needs.

\header{Image impact on answers}
The presence of images significantly influences answer characteristics and subsequent retrieval performance. Without images, users achieve higher retrieval scores (NDCG@1 of 0.185 vs. 0.079 in Original Query+QA), primarily because they provide more comprehensive textual responses to compensate for the absence of visual context. Conversely, with images, users tend to generate more specific answers that reference visual elements inaccessible to the text-based retrieval system.

\header{Image impact on reformulations}
Images substantially improve query reformulation quality, with NDCG@1 increasing from 0.091 to 0.204 in the Query-only condition. This improvement suggests that visual context helps users formulate more precise queries by providing concrete reference points and complementary contextual information for articulating search intent.

\subsection{Impact of images on answer consistency}
\begin{figure}[t]
\centering
\begin{subfigure}{.45\textwidth}
\centering
\includegraphics[width=\linewidth,height=48mm]{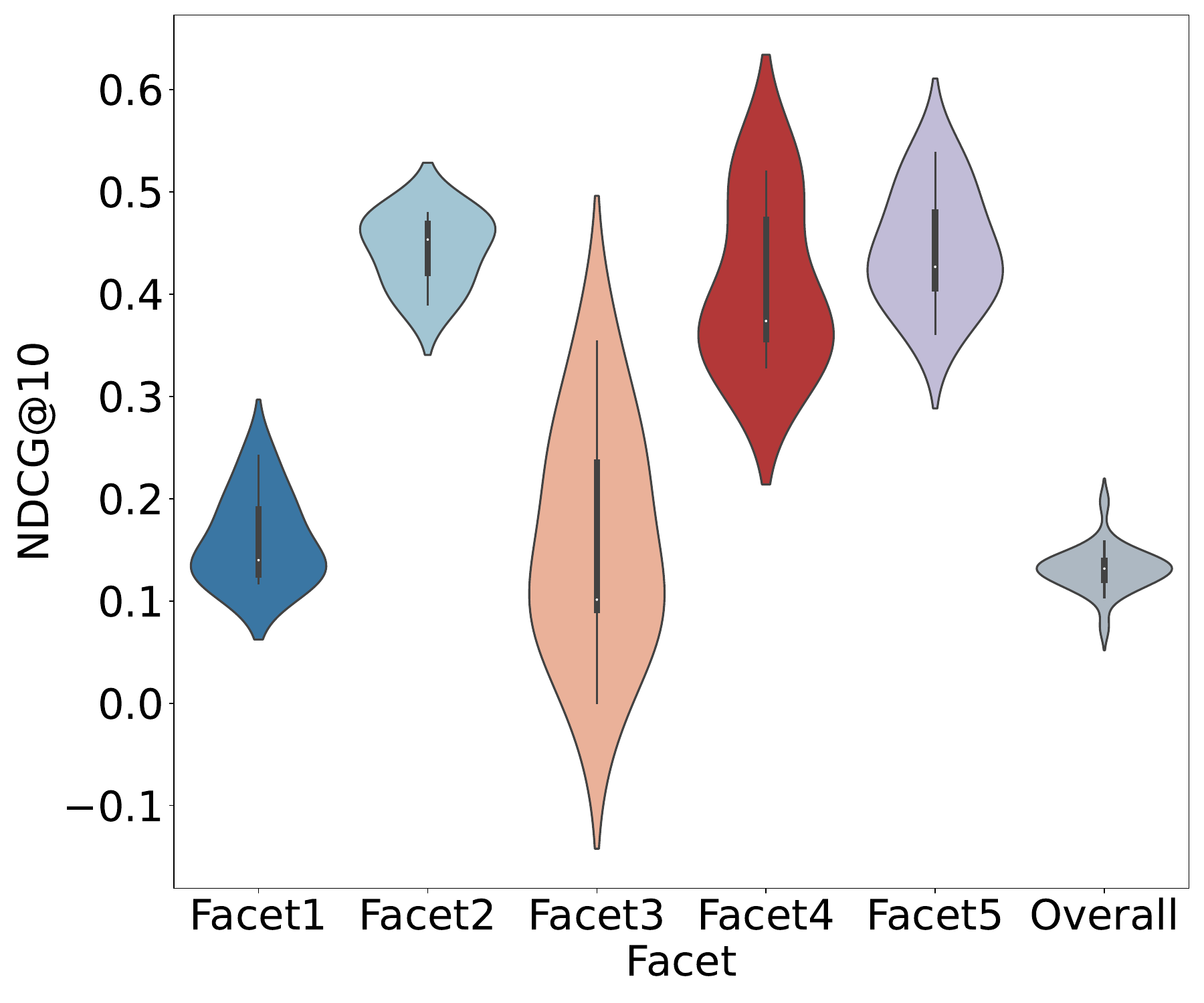}
\caption{With-image}
\label{fig:viol1}
\end{subfigure}%
\hfill
\begin{subfigure}{.45\textwidth}
\centering
\includegraphics[width=\linewidth,height=48mm]{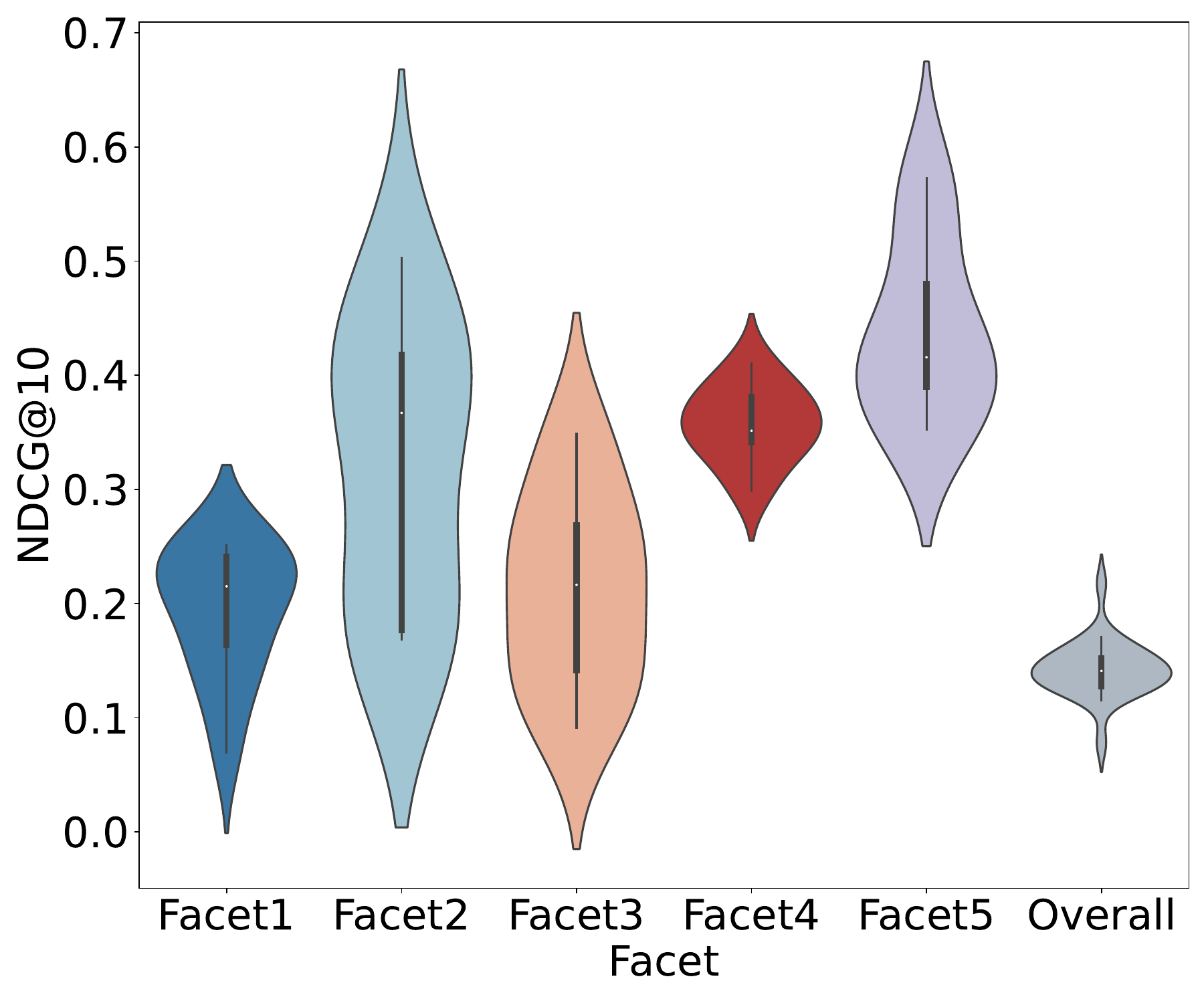}
\caption{Without-image}
\label{fig:viol2}
\end{subfigure}%
\hfill
\caption{Performance distribution across facets with different answers.}
\end{figure}
To understand how images influence user answer consistency, we analyzed performance distribution across five representative facets, selected to cover diverse query types. 
The results are shown in Figure \ref{fig:viol1} and \ref{fig:viol2}. 
The five selected facets are: \textit{Facet 1}: \textit{What specific health risks are associated with exposure to asbestos?}; \textit{Facet 2}: \textit{What home remedies are there for angular cheilitis?}; \textit{Facet 3}: \textit{What salary range does a land surveyor receive?}; \textit{Facet 4}: \textit{What were the results of the electoral college for the 2008 US presidential race?}; (5) \textit{Facet 5}: \textit{Is there a link between Barrett's Esophagus and cancer?}.

In most facets, images help reduce performance variability, leading to more consistent results across different answers. This effect is particularly evident in \textit{Facet 2}, where the performance distribution becomes more centered and narrower with images. For instance, when users can see actual images of angular cheilitis rather than relying on text descriptions alone, they generate more consistent answers focused on remedies specifically suited to the visible symptoms.

However, certain facets show contrasting effects. For \textit{Facet 3} and \textit{Facet 4}, which involve specific numeric information (salary ranges and electoral results), the performance distribution with images shows greater variability. This wider distribution indicates that for queries seeking precise, quantitative information, images can introduce distractions or unnecessary complexity, potentially leading to less consistent answers and retrieval outcomes.

In summary, while images could help reduce response variability and improve retrieval consistency in some facets, their overall impact is mixed. This suggests that the effectiveness of images in retrieval is facet-dependent and should be determined in specific search scenarios.

\subsection{Effect of reformulated queries on retrieval performance}
We present examples of the best- and worst-performing reformulated queries for a specific facet, comparing their performance across both with-image and without-image queries.

\header{Image-aligned reformulation}
Visual information appears to guide users toward more effective query reformulation when the images directly relate to key search aspects. For example, in Table \ref{tab:casestudy1}, when presented with images of ham radio antennas (row 3), users formulated queries that specifically referenced the visible antenna types, achieving better performance compared to the more general reformulation without visual context (row 4). This demonstrates how visual cues can help users incorporate relevant technical details into their queries.

\header{Precision in reformulation}
Our analysis reveals that precise, focused reformulations consistently outperform broader queries. In the angular cheilitis case (Table \ref{tab:casestudy2}, rows~1-2), a specific query about non-prescription remedies achieved substantially higher performance (NDCG: 0.479) compared to a more general reformulation (NDCG: 0.390). This precision advantage appears more when images help users identify specific aspects of their information needs.

\header{Facet-specific targeting}
Query reformulations that directly address user needs show superior performance. The Norway spruce example (Table \ref{tab:casestudy2}, row~3) illustrates this effect, where a focused query about planting and cultivation guidance (NDCG: 0.312) outperformed a broader, less targeted reformulation. Visual context appears to help users better align their reformulations with specific information needs rather than generating general-purpose queries.

These patterns suggest that visual enhancement can guide users toward more effective query reformulation strategies, particularly when images help users identify and incorporate specific, relevant details into their queries.

\begin{table}[]
\caption{Best and worst performed a reformulated query for each facet (with image).}
\footnotesize
\begin{tabular}{p{2.5cm}p{1cm}p{2.5cm}p{3cm}p{2.4cm}ll}
\toprule
Facet & Ori. Query & Reformulated Query & Clarifying Question & Image &NDCG & \\
\midrule
   What salary range does a land surveyor receive?     &        land surveyor       &           How much would a land surveyor typically earn.      & Are you interested in salary comparisons for land surveyors based on experience or geographic location?                    &   
   \multirow{2}{*}{\parbox[c]{0.2cm}{\centering\includegraphics[width=1.7cm,height=2.0cm]{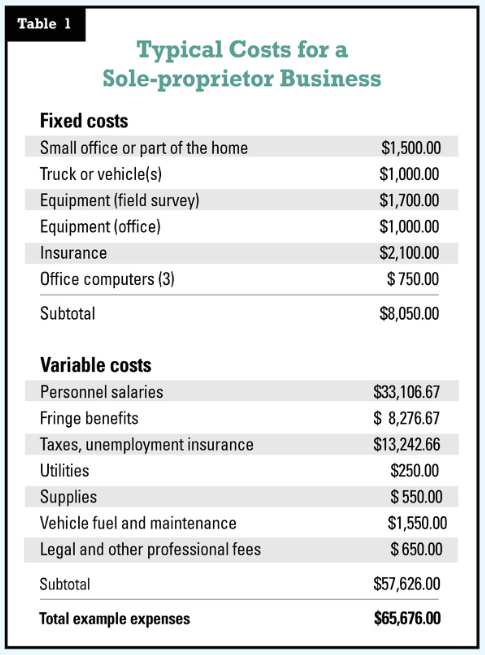}}}   &0.354      & \faThumbsUp      \\
 What salary range does a land surveyor receive?     &        land surveyor    &           What kind of remuneration does a land surveyor receive?     &  Are you interested in salary comparisons for land surveyors based on experience or geographic location?              &      &0.00  & \faThumbsDown
          \\
\midrule
  I need to learn more about ham  radio antennas. What types are there, and how are they used in communication?     &        ham radio        &       What types of ham radio antennas are there and how does a facilitate communication?             &   Are you looking for information on the different designs of ham radio antennas and their specific applications?                  &  
  \multirow{2}{*}{\raisebox{-1.5\totalheight}{\includegraphics[width=2.4cm,height=1.2cm]{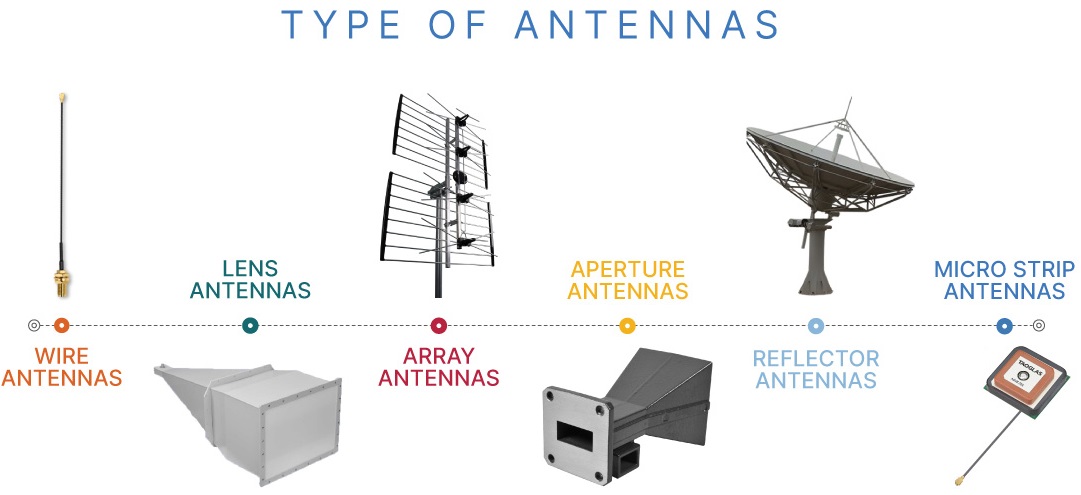}}}
  & 0.305         & \faThumbsUp        \\
  I need to learn more about ham  radio antennas. What types are there, and how are they used in communication?   &     ham radio           &       Where can I learn about amateur radio antennas and their applications?             &     Are you looking for information on the different designs of ham radio antennas and their specific applications?                 &   & 0.140  & \faThumbsDown\\          
 \bottomrule
\end{tabular}
\label{tab:casestudy1}
\end{table}

\begin{table}[]
\caption{Best and worst performed reformulated query for each facet (without image).}
\footnotesize
\begin{tabular}{p{4.0cm}p{1cm}p{2.5cm}p{4.3cm}ll}
\toprule
Facet & Ori. Query & Reformulated Query & Clarifying Question &NDCG & \\
\midrule
What home remedies are there for angular cheilitis?    &         angular cheilitis    &           non-prescription remedies for angular cheilitis     &  What types of home remedies are you interested in, or do you have specific concerns related to angular cheilitis?                   &0.479      & \faThumbsUp      \\
 What home remedies are there for angular cheilitis? &        angular cheilitis     &           What is the best home treatment for Cheilitis?     &  What types of home remedies are you interested in, or do you have specific concerns related to angular cheilitis?                 &0.390     & \faThumbsDown
          \\
\midrule
   I’m interested in gathering information on how to plant and cultivate Norway Spruce trees, including the best growth conditions and care tips.  &        norway spruce       &       Can you provide me with a guide to planting and cultivating Norway Spruce trees?            &   What particular aspects of planting and caring for Norway Spruce trees are you most focused on, such as growth conditions or maintenance tips?                  & 0.312         & \faThumbsUp        \\
   I’m interested in gathering information on how to plant and cultivate Norway Spruce trees, including the best growth conditions and care tips.   &norway spruce  &       Where can I grow Norwegian spruce trees successfully in the world?          &    What particular aspects of planting and caring for Norway Spruce trees are you most focused on, such as growth conditions or maintenance tips?                 & 0.178         & \faThumbsDown        \\
 \bottomrule
\end{tabular}
\label{tab:casestudy2}
\end{table}

\section{Analysis of Clarifying Answers and Reformulated Queries}

We conduct qualitative analysis to explore the impact of image availability on: responses to clarifying questions (Task 1) and query reformulation (Task 2).

\subsection{Task 1: Responses to clarifying questions}

Our analysis of Task 1 reveals a profound impact of visual cues on users' responses to clarifying questions. The presence of images consistently leads to more specific, detailed, and confident answers across various domains.

A striking finding is the marked increase in response specificity when images are available. For a meteorological topic, e.g., when asked about cloud types, a user with access to an image responds, ``I am seeing a stratus-shaped cloud''. This response not only demonstrates careful observation but also applies specific meteorological terminology. In contrast, without an image, another user answers the same question with a more general description: ``They are long, thin clouds''. This comparison vividly illustrates how visual cues can trigger the use of domain-specific language and more precise observations.

We observe a notable increase in the comprehensiveness of responses when images are absent, suggesting a compensatory strategy. For example, when asked about grilling techniques, an image-enhanced response is concise: ``How to enhance flavors?.'' Without an image, users tended to provide more elaborate answers, such as ``I would like you to present a comprehensive list of vegetable grilling suggestions to enhance the flavor of vegetables.'' This pattern indicates that in the absence of visual cues, users attempt to fill the information gap with more detailed verbal descriptions.

Furthermore, the presence of visual cues leads to an increased display of confidence in responses. When queried about dinosaur types, a user with image access stated, ``I am interested in a list of different kinds of dinosaurs.'' Without an image, responses are more tentative: ``I want a list of the types/names of dinosaurs including images to accompany.'' This difference in tone and directness underscores the role of visual information in reinforcing user confidence during the clarification process.

The results from Task 1 reveal that: First, images serve as cognitive anchors, enabling users to access and articulate domain-specific knowledge more precisely, particularly in technical and medical domains. Second, without images, users adopt a compensatory strategy of providing more comprehensive textual descriptions, effectively recreating missing visual context and leading to improved retrieval performance. Third, visual support increases response confidence, suggesting that images help validate user understanding and potentially streamline early-stage information exchange.

\subsection{Task 2: Query reformulation }

The analysis of Task 2 uncovers distinct patterns in how visual cues influence query reformulation. The presence of images leads to more focused, action-oriented, and specific query refinements.

In the domain of automotive information, we observed a clear trend toward increased specificity in image-enhanced reformulations. An initial query about ``Car dashboard symbols meaning'' was refined to ``Could you explain what the airbag indicator light means on a car dashboard?'' when an image was present. Without visual cues, the reformulation remained broad: ``what is the meaning of car dashboard symbols?''. This demonstrates how visual information enables users to pinpoint specific aspects of their information needs during the reformulation process.

Visual cues also prompt a shift towards more action-oriented queries. For, an initial query about ``grilling'' it evolved into ``Vegetable grilling techniques to enhance flavor and presentation'' with image assistance. Without visual cues, users tended towards more comprehensive information gathering: ``I want some specific techniques when grilling that will help the taste, without ruining the visual appeal of the dish''. This pattern suggests that visual information encourages users to focus on practical applications and specific actions.

The influence of visual cues on query scope is particularly noteworthy. In a query about Norway spruce trees, image-enhanced reformulation led to a focused request: ``Can you provide me with a guide to planting and cultivating Norway Spruce trees?''. Without images, users sought broader information: ``What are the best growth conditions and maintenance tips for the planting and caring of Norway spruce trees?''. This divergence in query scope highlights how visual cues can shape the breadth and depth of information-seeking behavior.

Task 2 findings highlight four key effects of visual cues on query reformulation strategies. First, images guide users toward more specific and focused queries, potentially enabling more efficient search through precise scope definition. Second, visual information promotes action-oriented reformulations, suggesting that images help users conceptualize practical applications of their information needs. Third, users demonstrate compensatory behavior without images, but unlike Task 1's verbose descriptions, this manifests as a broader query scope. Finally, visual support increases directness in query formulation, indicating that images enhance user confidence in articulating specific information needs.

\header{Summary}
Our analysis reveals the selective effectiveness of visual enhancement in \ac{CS}: the impact of image availability varies with query type and domain.
Technical and product-related queries showed substantially different patterns of specificity and detail between with- and without-image conditions, highlighting domains where visual support plays a crucial role. However, for general or conceptual questions, these differences were notably less pronounced, suggesting that visual enhancement provides selective rather than universal benefits. This domain and query-dependent pattern has important implications for the design of multimodal search systems, indicating that visual enhancement strategies should be adaptively deployed based on query characteristics rather than uniformly applied across all search scenarios.

\section{Discussion and Implications}
\subsection{Discussion}

In \ref{rq1} we ask \textit{``How do images influence users’ answers to clarifying questions in \ac{CS}?''}. Our findings reveal several key insights about the role of visual elements in \ac{CS} interactions. First, despite the common assumption that visual aids enhance clarity, we found that text-only questions were perceived as marginally clearer. This unexpected finding suggests that integrating visual information may introduce additional complexity to the clarification process. However, this increased complexity does not necessarily detract from the overall user experience. Users consistently demonstrated a strong preference for multimodal questions (61.1\%). The minimal increase in task completion time suggests efficient integration of visual information, indicating that any cognitive overhead from processing images was effectively balanced by their contextual benefits. This efficient integration of visual information aligns with dual coding theory~\citep{paivio1991dual}, which suggests that parallel processing of visual and verbal information can be cognitively efficient.

The varying impact of images across different types of search tasks is particularly notable. The dominance of physical object queries (50\%) among the scenarios in which images proved the most useful suggests that visual elements are particularly valuable when clarifying tangible or spatial characteristics. This finding extends previous research on visual information in search tasks~\citep{Yuan2024Asking} by identifying specific contexts where visual enhancement is most beneficial. 
Intriguingly, our results shed new light on how visual elements influence users with different levels of domain expertise. 
We observed that user background knowledge had less impact on their performance with multimodal questions, suggesting that image-enhanced questions provide enough context to enable users with different background knowledge equally.
This finding contributes to our understanding of expertise-based differences in search behavior~\citep{DBLP:conf/chiir/OBrienCKB22} and highlights the potential of visual elements to create more adaptive and inclusive search experiences.

While users report positive experiences with image-enhanced interactions in certain contexts, this did not consistently translate to improved performance. Higher user satisfaction with images did not necessarily correlate with better retrieval outcomes of their answers. For clarifying question answers, despite the positive reception of images by users, responses without images achieved better retrieval scores (NDCG@1 0.185 vs.\ 0.079), probably because users provided more comprehensive textual information when compensating for the lack of visual context, and the fact that our retrieval model did not include visual embeddings into account. 

In \ref{rq2} we examine ``\textit{What effect do images have on query reformulation in \ac{CS}?}''. We show that images play a distinct and complex role in query reformulation. While all users significantly expanded their queries (from 3.13 to 12.46 words on average), those with access to images produced slightly more concise reformulations. This suggests that visual information helps users focus their information needs more precisely, potentially leading to more efficient search strategies. This finding extends previous work on query reformulation~\citep{DBLP:conf/cikm/HuangE09} by demonstrating how visual elements can help streamline query modification.
However, the small differences in usefulness and satisfaction suggest that images are not always perceived as universally beneficial during query reformulation, indicating that images can either enhance or distract users, depending on the task at hand. For some participants, visual cues may provide the additional context needed to clarify their initial queries, while for others, this additional information may introduce unnecessary complexity or cognitive load.

Interestingly, the low frequency of direct image usage in questions where images are not explicitly required suggests that users tend to rely on visual information only when they perceive it to be necessary. This indicates that users favor visual content when they feel it adds substantial value.
Participants were less likely to refer to their explicitly stated information needs when images were present. Images provide an implicit form of context that enables users to internalize and restructure their information needs without needing to explicitly recall them. This phenomenon aligns with \citet{DBLP:conf/cogsci/LuKR20}'s theory of cognitive offloading, where users rely on external tools, such as images, to simplify the mental processes required for query reformulation.

In \ref{rq3} we investigate \textit{``When are images useful in query clarification?''}.
Our investigation of image utility in \ac{CS}, combining situational analysis and thematic coding of user responses, reveals three critical dimensions that determine the effectiveness of visual elements in clarifying questions:

First, query characteristics emerge as a primary determinant, supported by both our situational and thematic analyses. Images demonstrate clear utility for queries with inherent visual attributes, particularly those involving physical objects (approximately 50\% of beneficial cases) and procedural information (14-19\%). Our thematic analysis reveals how this utility manifests through contextual support (28.4\%) and cognitive facilitation (24.3\%), especially when visual elements align closely with query intent. However, for abstract queries, images can create cognitive overload (13.5\%) or become irrelevant to task goals (10.8\%), suggesting that visual enhancement decisions should carefully consider query type~(Section~\ref{sec:img-usefulness}). This finding extends previous work on multimodal search clarification~\cite{Yuan2024Asking} by specifically identifying which query types benefit most from visual enhancement in conversational contexts.

Second, the type of search task significantly moderates image effectiveness, influencing how users use visual information. While direct question answering benefits from images through creative stimulation (21.6\%) and increased engagement (16.2\%), query reformulation shows a more complex relationship with visual elements, sometimes leading to redundancy (8.1\%). This task dependency suggests that mixed-initiative systems need to consider not only whether to include images, but when in the conversation they would be most beneficial.

Third, user expertise plays a crucial role in image utility in our thematic findings. Expert users demonstrate particularly effective use of visual information when present, showing an enhanced ability to integrate visual cues into their search strategies. Although few users (9.5\%) expressed a preference for text-only interactions, our analysis shows that the level of expertise significantly influences how effectively visual information is used.

\subsection{Implications and limitations}
Our findings reveal requirements for designing effective multimodal \ac{CS} systems. First, systems need to implement adaptive visual enhancement strategies based on task characteristics. Our results demonstrate distinct patterns of image utility between answering clarifying questions and query reformulation tasks, suggesting the need for mechanisms that dynamically assess when to present visual information based on the current search stage and task requirements.

Second, user expertise significantly influences image effectiveness, indicating the need for expertise-aware visual support. Systems should adapt their visual strategy based on user expertise signals, potentially varying both the complexity and quantity of visual information. This could involve developing presentation strategies for novice users who benefit from basic visual context versus expert users who can leverage more complex visual information.

Third, query characteristics should drive image deployment decisions. The clear correlation between query type and image utility suggests implementing automatic assessment of query visual dependency. Systems should prioritize visual enhancement for queries with inherent visual attributes (e.g., physical objects and processes) while limiting visual elements for abstract queries.

However, several limitations should be considered when interpreting these implications. Our study used a curated subset of ClariQ dataset topics, which may not fully represent the diversity of real-world search scenarios. The controlled nature of our user study, while enabling systematic analysis, used pre-defined queries rather than natural \ac{CS} interactions. Additionally, while we employed systematic image selection, we did not explore how different image characteristics might affect search interaction. Given the complex nature of visual elements, a deeper study of the images exhibiting specific visual characteristics can complement our work. Our work, however, provides important insights into this task and the role of images on user's perceptions that can inform future studies.

\section{Conclusion}

In this study, we have investigated the effect of images on \ac{CS}, through a user study examining their role in answering clarifying questions and query reformulation. Our findings reveal that while images generally enhance the search experience, their impact varies significantly based on search task, user expertise, and question characteristics. Across tasks, images are found to enhance clarity and usefulness in specific contexts, particularly when users seek information about physical objects or processes. However, even though images can help users refine their queries and improve search efficiency, their utility is task-dependent. Visual aids are most effective when clarifying tangible items and complex procedures, but less so for abstract concepts and context-related queries.

Our work contributes to the growing understanding of multimodal interaction in \ac{CS} by providing empirical evidence of when and how images can improve search outcomes. Future work should explore how to optimize the integration of images into real-world search systems and investigate the long-term effects of multimodal interfaces on user satisfaction and performance.

\begin{acks}
This research was supported by 
Huawei Finland,
the Dutch Research Council (NWO), under project numbers 024.004.022, NWA.1389.20.\-183, and KICH3.LTP.20.006,
and the European Union's Horizon Europe program under grant agreement No 101070212.
All content represents the opinion of the authors, which is not necessarily shared or endorsed by their respective employers and/or sponsors.
\end{acks}

\bibliographystyle{ACM-Reference-Format}
\bibliography{references}


\begin{thebibliography}{54}


\ifx \showCODEN    \undefined \def \showCODEN     #1{\unskip}     \fi
\ifx \showDOI      \undefined \def \showDOI       #1{#1}\fi
\ifx \showISBNx    \undefined \def \showISBNx     #1{\unskip}     \fi
\ifx \showISBNxiii \undefined \def \showISBNxiii  #1{\unskip}     \fi
\ifx \showISSN     \undefined \def \showISSN      #1{\unskip}     \fi
\ifx \showLCCN     \undefined \def \showLCCN      #1{\unskip}     \fi
\ifx \shownote     \undefined \def \shownote      #1{#1}          \fi
\ifx \showarticletitle \undefined \def \showarticletitle #1{#1}   \fi
\ifx \showURL      \undefined \def \showURL       {\relax}        \fi
\providecommand\bibfield[2]{#2}
\providecommand\bibinfo[2]{#2}
\providecommand\natexlab[1]{#1}
\providecommand\showeprint[2][]{arXiv:#2}

\bibitem[Aliannejadi et~al\mbox{.}(2021a)]%
        {Aliannejadi2021AnalysingMI}
\bibfield{author}{\bibinfo{person}{Mohammad Aliannejadi}, \bibinfo{person}{Leif Azzopardi}, \bibinfo{person}{Hamed Zamani}, \bibinfo{person}{Evangelos Kanoulas}, \bibinfo{person}{Paul Thomas}, {and} \bibinfo{person}{Nick Craswell}.} \bibinfo{year}{2021}\natexlab{a}.
\newblock \showarticletitle{Analysing Mixed Initiatives and Search Strategies during Conversational Search}. In \bibinfo{booktitle}{\emph{{CIKM} '21: The 30th {ACM} International Conference on Information and Knowledge Management, Virtual Event, Queensland, Australia, November 1 - 5, 2021}}, \bibfield{editor}{\bibinfo{person}{Gianluca Demartini}, \bibinfo{person}{Guido Zuccon}, \bibinfo{person}{J.~Shane Culpepper}, \bibinfo{person}{Zi~Huang}, {and} \bibinfo{person}{Hanghang Tong}} (Eds.). \bibinfo{publisher}{{ACM}}, \bibinfo{pages}{16--26}.
\newblock
\urldef\tempurl%
\url{https://doi.org/10.1145/3459637.3482231}
\showDOI{\tempurl}


\bibitem[Aliannejadi et~al\mbox{.}(2020)]%
        {Aliannejadi2020ConvAI3GC}
\bibfield{author}{\bibinfo{person}{Mohammad Aliannejadi}, \bibinfo{person}{Julia Kiseleva}, \bibinfo{person}{Aleksandr Chuklin}, \bibinfo{person}{Jeff Dalton}, {and} \bibinfo{person}{Mikhail Burtsev}.} \bibinfo{year}{2020}\natexlab{}.
\newblock \showarticletitle{ConvAI3: Generating Clarifying Questions for Open-Domain Dialogue Systems (ClariQ)}.
\newblock \bibinfo{journal}{\emph{CoRR}}  \bibinfo{volume}{abs/2009.11352} (\bibinfo{year}{2020}).
\newblock
\showeprint[arXiv]{2009.11352}
\urldef\tempurl%
\url{https://arxiv.org/abs/2009.11352}
\showURL{%
\tempurl}


\bibitem[Aliannejadi et~al\mbox{.}(2021b)]%
        {aliannejadi-etal-2021-building}
\bibfield{author}{\bibinfo{person}{Mohammad Aliannejadi}, \bibinfo{person}{Julia Kiseleva}, \bibinfo{person}{Aleksandr Chuklin}, \bibinfo{person}{Jeff Dalton}, {and} \bibinfo{person}{Mikhail Burtsev}.} \bibinfo{year}{2021}\natexlab{b}.
\newblock \showarticletitle{Building and Evaluating Open-Domain Dialogue Corpora with Clarifying Questions}. In \bibinfo{booktitle}{\emph{Proceedings of the 2021 Conference on Empirical Methods in Natural Language Processing, {EMNLP} 2021, Virtual Event / Punta Cana, Dominican Republic, 7-11 November, 2021}}, \bibfield{editor}{\bibinfo{person}{Marie{-}Francine Moens}, \bibinfo{person}{Xuanjing Huang}, \bibinfo{person}{Lucia Specia}, {and} \bibinfo{person}{Scott~Wen{-}tau Yih}} (Eds.). \bibinfo{publisher}{Association for Computational Linguistics}, \bibinfo{pages}{4473--4484}.
\newblock
\urldef\tempurl%
\url{https://doi.org/10.18653/V1/2021.EMNLP-MAIN.367}
\showDOI{\tempurl}


\bibitem[Aliannejadi et~al\mbox{.}(2019)]%
        {AliannejadiSigir19}
\bibfield{author}{\bibinfo{person}{Mohammad Aliannejadi}, \bibinfo{person}{Hamed Zamani}, \bibinfo{person}{Fabio Crestani}, {and} \bibinfo{person}{W.~Bruce Croft}.} \bibinfo{year}{2019}\natexlab{}.
\newblock \showarticletitle{Asking Clarifying Questions in Open-Domain Information-Seeking Conversations}. In \bibinfo{booktitle}{\emph{Proceedings of the 42nd International {ACM} {SIGIR} Conference on Research and Development in Information Retrieval, {SIGIR} 2019, Paris, France, July 21-25, 2019}}, \bibfield{editor}{\bibinfo{person}{Benjamin Piwowarski}, \bibinfo{person}{Max Chevalier}, \bibinfo{person}{{\'{E}}ric Gaussier}, \bibinfo{person}{Yoelle Maarek}, \bibinfo{person}{Jian{-}Yun Nie}, {and} \bibinfo{person}{Falk Scholer}} (Eds.). \bibinfo{publisher}{{ACM}}, \bibinfo{pages}{475--484}.
\newblock
\urldef\tempurl%
\url{https://doi.org/10.1145/3331184.3331265}
\showDOI{\tempurl}


\bibitem[Allen et~al\mbox{.}(1999)]%
        {Allen1999MixedinitiativeI}
\bibfield{author}{\bibinfo{person}{James Allen}, \bibinfo{person}{Curry~I. Guinn}, {and} \bibinfo{person}{Eric Horvitz}.} \bibinfo{year}{1999}\natexlab{}.
\newblock \showarticletitle{Mixed-initiative Interaction}.
\newblock \bibinfo{journal}{\emph{IEEE Intelligent Systems \& Their Applications}}  \bibinfo{volume}{14} (\bibinfo{year}{1999}).
\newblock


\bibitem[Avula et~al\mbox{.}(2022)]%
        {DBLP:journals/pacmhci/AvulaCA22}
\bibfield{author}{\bibinfo{person}{Sandeep Avula}, \bibinfo{person}{Bogeum Choi}, {and} \bibinfo{person}{Jaime Arguello}.} \bibinfo{year}{2022}\natexlab{}.
\newblock \showarticletitle{The Effects of System Initiative during Conversational Collaborative Search}.
\newblock \bibinfo{journal}{\emph{Proc. {ACM} Hum. Comput. Interact.}} \bibinfo{volume}{6}, \bibinfo{number}{{CSCW1}} (\bibinfo{year}{2022}), \bibinfo{pages}{66:1--66:30}.
\newblock
\urldef\tempurl%
\url{https://doi.org/10.1145/3512913}
\showDOI{\tempurl}


\bibitem[Azzopardi et~al\mbox{.}(2022)]%
        {Azzopardi2022Towards}
\bibfield{author}{\bibinfo{person}{Leif Azzopardi}, \bibinfo{person}{Mohammad Aliannejadi}, {and} \bibinfo{person}{Evangelos Kanoulas}.} \bibinfo{year}{2022}\natexlab{}.
\newblock \showarticletitle{Towards Building Economic Models of Conversational Search}. In \bibinfo{booktitle}{\emph{{ECIR} {(2)}}} \emph{(\bibinfo{series}{Lecture Notes in Computer Science}, Vol.~\bibinfo{volume}{13186})}. \bibinfo{publisher}{Springer}, \bibinfo{pages}{31--38}.
\newblock


\bibitem[Bobek and Tversky(2014)]%
        {bobek-2016-creating}
\bibfield{author}{\bibinfo{person}{Eliza Bobek} {and} \bibinfo{person}{Barbara Tversky}.} \bibinfo{year}{2014}\natexlab{}.
\newblock \showarticletitle{Creating Visual Explanations Improves Learning}.
\newblock  (\bibinfo{year}{2014}).
\newblock
\urldef\tempurl%
\url{https://escholarship.org/uc/item/3gh2z11f}
\showURL{%
\tempurl}


\bibitem[Braslavski et~al\mbox{.}(2017)]%
        {Braslavski2017WhatDY}
\bibfield{author}{\bibinfo{person}{Pavel Braslavski}, \bibinfo{person}{Denis Savenkov}, \bibinfo{person}{Eugene Agichtein}, {and} \bibinfo{person}{Alina Dubatovka}.} \bibinfo{year}{2017}\natexlab{}.
\newblock \showarticletitle{What Do You Mean Exactly?: Analyzing Clarification Questions in {CQA}}. In \bibinfo{booktitle}{\emph{Proceedings of the 2017 Conference on Conference Human Information Interaction and Retrieval, {CHIIR} 2017, Oslo, Norway, March 7-11, 2017}}, \bibfield{editor}{\bibinfo{person}{Ragnar Nordlie}, \bibinfo{person}{Nils Pharo}, \bibinfo{person}{Luanne Freund}, \bibinfo{person}{Birger Larsen}, {and} \bibinfo{person}{Dan Russel}} (Eds.). \bibinfo{publisher}{{ACM}}, \bibinfo{pages}{345--348}.
\newblock
\urldef\tempurl%
\url{https://doi.org/10.1145/3020165.3022149}
\showDOI{\tempurl}


\bibitem[Chang et~al\mbox{.}(2022)]%
        {Chang2021WebQAMA}
\bibfield{author}{\bibinfo{person}{Yingshan Chang}, \bibinfo{person}{Guihong Cao}, \bibinfo{person}{Mridu Narang}, \bibinfo{person}{Jianfeng Gao}, \bibinfo{person}{Hisami Suzuki}, {and} \bibinfo{person}{Yonatan Bisk}.} \bibinfo{year}{2022}\natexlab{}.
\newblock \showarticletitle{WebQA: Multihop and Multimodal {QA}}. In \bibinfo{booktitle}{\emph{{IEEE/CVF} Conference on Computer Vision and Pattern Recognition, {CVPR} 2022, New Orleans, LA, USA, June 18-24, 2022}}. \bibinfo{publisher}{{IEEE}}, \bibinfo{pages}{16474--16483}.
\newblock
\urldef\tempurl%
\url{https://doi.org/10.1109/CVPR52688.2022.01600}
\showDOI{\tempurl}


\bibitem[Clarke et~al\mbox{.}(2008)]%
        {clarke-2008-novelty}
\bibfield{author}{\bibinfo{person}{Charles L.~A. Clarke}, \bibinfo{person}{Maheedhar Kolla}, \bibinfo{person}{Gordon~V. Cormack}, \bibinfo{person}{Olga Vechtomova}, \bibinfo{person}{Azin Ashkan}, \bibinfo{person}{Stefan B{\"{u}}ttcher}, {and} \bibinfo{person}{Ian MacKinnon}.} \bibinfo{year}{2008}\natexlab{}.
\newblock \showarticletitle{Novelty and diversity in information retrieval evaluation}. In \bibinfo{booktitle}{\emph{Proceedings of the 31st Annual International {ACM} {SIGIR} Conference on Research and Development in Information Retrieval, {SIGIR} 2008, Singapore, July 20-24, 2008}}, \bibfield{editor}{\bibinfo{person}{Sung{-}Hyon Myaeng}, \bibinfo{person}{Douglas~W. Oard}, \bibinfo{person}{Fabrizio Sebastiani}, \bibinfo{person}{Tat{-}Seng Chua}, {and} \bibinfo{person}{Mun{-}Kew Leong}} (Eds.). \bibinfo{publisher}{{ACM}}, \bibinfo{pages}{659--666}.
\newblock
\urldef\tempurl%
\url{https://doi.org/10.1145/1390334.1390446}
\showDOI{\tempurl}


\bibitem[Collins{-}Thompson et~al\mbox{.}(2014)]%
        {Collins2014Web}
\bibfield{author}{\bibinfo{person}{Kevyn Collins{-}Thompson}, \bibinfo{person}{Craig Macdonald}, \bibinfo{person}{Paul~N. Bennett}, \bibinfo{person}{Fernando Diaz}, {and} \bibinfo{person}{Ellen~M. Voorhees}.} \bibinfo{year}{2014}\natexlab{}.
\newblock \showarticletitle{{TREC} 2014 Web Track Overview}. In \bibinfo{booktitle}{\emph{{TREC}}} \emph{(\bibinfo{series}{{NIST} Special Publication}, Vol.~\bibinfo{volume}{500-308})}. \bibinfo{publisher}{National Institute of Standards and Technology {(NIST)}}.
\newblock


\bibitem[Dagan et~al\mbox{.}(2021)]%
        {DBLP:conf/sigir/DaganGN21}
\bibfield{author}{\bibinfo{person}{Arnon Dagan}, \bibinfo{person}{Ido Guy}, {and} \bibinfo{person}{Slava Novgorodov}.} \bibinfo{year}{2021}\natexlab{}.
\newblock \showarticletitle{An Image is Worth a Thousand Terms? Analysis of Visual E-Commerce Search}. In \bibinfo{booktitle}{\emph{{SIGIR} '21: The 44th International {ACM} {SIGIR} Conference on Research and Development in Information Retrieval, Virtual Event, Canada, July 11-15, 2021}}, \bibfield{editor}{\bibinfo{person}{Fernando Diaz}, \bibinfo{person}{Chirag Shah}, \bibinfo{person}{Torsten Suel}, \bibinfo{person}{Pablo Castells}, \bibinfo{person}{Rosie Jones}, {and} \bibinfo{person}{Tetsuya Sakai}} (Eds.). \bibinfo{publisher}{{ACM}}, \bibinfo{pages}{102--112}.
\newblock
\urldef\tempurl%
\url{https://doi.org/10.1145/3404835.3462950}
\showDOI{\tempurl}


\bibitem[Di et~al\mbox{.}(2014)]%
        {DBLP:conf/wsdm/DiSPB14}
\bibfield{author}{\bibinfo{person}{Wei Di}, \bibinfo{person}{Neel Sundaresan}, \bibinfo{person}{Robinson Piramuthu}, {and} \bibinfo{person}{Anurag Bhardwaj}.} \bibinfo{year}{2014}\natexlab{}.
\newblock \showarticletitle{Is a Picture Really Worth a Thousand Words?: -- On the Role of Images in E-commerce}. In \bibinfo{booktitle}{\emph{Seventh {ACM} International Conference on Web Search and Data Mining, {WSDM} 2014, New York, NY, USA, February 24-28, 2014}}, \bibfield{editor}{\bibinfo{person}{Ben Carterette}, \bibinfo{person}{Fernando Diaz}, \bibinfo{person}{Carlos Castillo}, {and} \bibinfo{person}{Donald Metzler}} (Eds.). \bibinfo{publisher}{{ACM}}, \bibinfo{pages}{633--642}.
\newblock
\urldef\tempurl%
\url{https://doi.org/10.1145/2556195.2556226}
\showDOI{\tempurl}


\bibitem[Fichter and Jonas(2008)]%
        {fichter2008image}
\bibfield{author}{\bibinfo{person}{Christian Fichter} {and} \bibinfo{person}{Klaus Jonas}.} \bibinfo{year}{2008}\natexlab{}.
\newblock \showarticletitle{Image Effects of Newspapers: How Brand Images Change Consumers’ Product Ratings}.
\newblock \bibinfo{journal}{\emph{Zeitschrift f{\"u}r Psychologie/Journal of Psychology}} \bibinfo{volume}{216}, \bibinfo{number}{4} (\bibinfo{year}{2008}), \bibinfo{pages}{226--234}.
\newblock


\bibitem[Hashemi et~al\mbox{.}(2020)]%
        {Hashemi2020GuidedTL}
\bibfield{author}{\bibinfo{person}{Helia Hashemi}, \bibinfo{person}{Hamed Zamani}, {and} \bibinfo{person}{W.~Bruce Croft}.} \bibinfo{year}{2020}\natexlab{}.
\newblock \showarticletitle{Guided Transformer: Leveraging Multiple External Sources for Representation Learning in Conversational Search}. In \bibinfo{booktitle}{\emph{Proceedings of the 43rd International {ACM} {SIGIR} conference on research and development in Information Retrieval, {SIGIR} 2020, Virtual Event, China, July 25-30, 2020}}, \bibfield{editor}{\bibinfo{person}{Jimmy~X. Huang}, \bibinfo{person}{Yi~Chang}, \bibinfo{person}{Xueqi Cheng}, \bibinfo{person}{Jaap Kamps}, \bibinfo{person}{Vanessa Murdock}, \bibinfo{person}{Ji{-}Rong Wen}, {and} \bibinfo{person}{Yiqun Liu}} (Eds.). \bibinfo{publisher}{{ACM}}, \bibinfo{pages}{1131--1140}.
\newblock
\urldef\tempurl%
\url{https://doi.org/10.1145/3397271.3401061}
\showDOI{\tempurl}


\bibitem[Hearst(1999)]%
        {Hearst1999TrendsC}
\bibfield{author}{\bibinfo{person}{Marti~A. Hearst}.} \bibinfo{year}{1999}\natexlab{}.
\newblock \showarticletitle{Trends {\&} Controversies: Mixed-initiative interaction}.
\newblock \bibinfo{journal}{\emph{{IEEE} Intell. Syst.}} \bibinfo{volume}{14}, \bibinfo{number}{5} (\bibinfo{year}{1999}), \bibinfo{pages}{14--23}.
\newblock
\urldef\tempurl%
\url{https://doi.org/10.1109/5254.796083}
\showDOI{\tempurl}


\bibitem[Houts et~al\mbox{.}(2006)]%
        {HOUTS2006173}
\bibfield{author}{\bibinfo{person}{Peter~S. Houts}, \bibinfo{person}{Cecilia~C. Doak}, \bibinfo{person}{Leonard~G. Doak}, {and} \bibinfo{person}{Matthew~J. Loscalzo}.} \bibinfo{year}{2006}\natexlab{}.
\newblock \showarticletitle{The Role of Pictures in Improving Health Communication: A Review of Research on Attention, Comprehension, Recall, and Adherence}.
\newblock \bibinfo{journal}{\emph{Patient Education and Counseling}} \bibinfo{volume}{61}, \bibinfo{number}{2} (\bibinfo{year}{2006}), \bibinfo{pages}{173--190}.
\newblock
\showISSN{0738-3991}
\urldef\tempurl%
\url{https://doi.org/10.1016/j.pec.2005.05.004}
\showDOI{\tempurl}


\bibitem[Huang and Efthimiadis(2009)]%
        {DBLP:conf/cikm/HuangE09}
\bibfield{author}{\bibinfo{person}{Jeff Huang} {and} \bibinfo{person}{Efthimis~N. Efthimiadis}.} \bibinfo{year}{2009}\natexlab{}.
\newblock \showarticletitle{Analyzing and Evaluating Query Reformulation Strategies in Web Search Logs}. In \bibinfo{booktitle}{\emph{Proceedings of the 18th {ACM} Conference on Information and Knowledge Management, {CIKM} 2009, Hong Kong, China, November 2-6, 2009}}, \bibfield{editor}{\bibinfo{person}{David~Wai{-}Lok Cheung}, \bibinfo{person}{Il{-}Yeol Song}, \bibinfo{person}{Wesley~W. Chu}, \bibinfo{person}{Xiaohua Hu}, {and} \bibinfo{person}{Jimmy Lin}} (Eds.). \bibinfo{publisher}{{ACM}}, \bibinfo{pages}{77--86}.
\newblock
\urldef\tempurl%
\url{https://doi.org/10.1145/1645953.1645966}
\showDOI{\tempurl}


\bibitem[Kelly(2009)]%
        {DBLP:journals/ftir/Kelly09}
\bibfield{author}{\bibinfo{person}{Diane Kelly}.} \bibinfo{year}{2009}\natexlab{}.
\newblock \showarticletitle{Methods for Evaluating Interactive Information Retrieval Systems with Users}.
\newblock \bibinfo{journal}{\emph{Found. Trends Inf. Retr.}} \bibinfo{volume}{3}, \bibinfo{number}{1-2} (\bibinfo{year}{2009}), \bibinfo{pages}{1--224}.
\newblock
\urldef\tempurl%
\url{https://doi.org/10.1561/1500000012}
\showDOI{\tempurl}


\bibitem[Kiesel et~al\mbox{.}(2018)]%
        {Kiesel2018TowardVQ}
\bibfield{author}{\bibinfo{person}{Johannes Kiesel}, \bibinfo{person}{Arefeh Bahrami}, \bibinfo{person}{Benno Stein}, \bibinfo{person}{Avishek Anand}, {and} \bibinfo{person}{Matthias Hagen}.} \bibinfo{year}{2018}\natexlab{}.
\newblock \showarticletitle{Toward Voice Query Clarification}. In \bibinfo{booktitle}{\emph{The 41st International {ACM} {SIGIR} Conference on Research {\&} Development in Information Retrieval, {SIGIR} 2018, Ann Arbor, MI, USA, July 08-12, 2018}}, \bibfield{editor}{\bibinfo{person}{Kevyn Collins{-}Thompson}, \bibinfo{person}{Qiaozhu Mei}, \bibinfo{person}{Brian~D. Davison}, \bibinfo{person}{Yiqun Liu}, {and} \bibinfo{person}{Emine Yilmaz}} (Eds.). \bibinfo{publisher}{{ACM}}, \bibinfo{pages}{1257--1260}.
\newblock
\urldef\tempurl%
\url{https://doi.org/10.1145/3209978.3210160}
\showDOI{\tempurl}


\bibitem[Krasakis et~al\mbox{.}(2020)]%
        {Krasakis2020AnalysingTE}
\bibfield{author}{\bibinfo{person}{Antonios~Minas Krasakis}, \bibinfo{person}{Mohammad Aliannejadi}, \bibinfo{person}{Nikos Voskarides}, {and} \bibinfo{person}{Evangelos Kanoulas}.} \bibinfo{year}{2020}\natexlab{}.
\newblock \showarticletitle{Analysing the Effect of Clarifying Questions on Document Ranking in Conversational Search}.
\newblock  (\bibinfo{year}{2020}), \bibinfo{pages}{129--132}.
\newblock
\urldef\tempurl%
\url{https://doi.org/10.1145/3409256.3409817}
\showDOI{\tempurl}


\bibitem[Li and Xie(2020)]%
        {li2020picture}
\bibfield{author}{\bibinfo{person}{Yiyi Li} {and} \bibinfo{person}{Ying Xie}.} \bibinfo{year}{2020}\natexlab{}.
\newblock \showarticletitle{Is a Picture Worth a Thousand Words? An Empirical Study of Image Content and Social Media Engagement}.
\newblock \bibinfo{journal}{\emph{Journal of Marketing Research}} \bibinfo{volume}{57}, \bibinfo{number}{1} (\bibinfo{year}{2020}), \bibinfo{pages}{1--19}.
\newblock


\bibitem[Lu et~al\mbox{.}(2020)]%
        {DBLP:conf/cogsci/LuKR20}
\bibfield{author}{\bibinfo{person}{Xinyi Lu}, \bibinfo{person}{Megan Kelly}, {and} \bibinfo{person}{Evan~F. Risko}.} \bibinfo{year}{2020}\natexlab{}.
\newblock \showarticletitle{Cognitive Offloading Increases False Recall}. In \bibinfo{booktitle}{\emph{Proceedings of the 42th Annual Meeting of the Cognitive Science Society - Developing a Mind: Learning in Humans, Animals, and Machines, CogSci 2020, virtual, July 29 - August 1, 2020}}, \bibfield{editor}{\bibinfo{person}{Stephanie Denison}, \bibinfo{person}{Michael~L. Mack}, \bibinfo{person}{Yang Xu}, {and} \bibinfo{person}{Blair~C. Armstrong}} (Eds.). \bibinfo{publisher}{cognitivesciencesociety.org}.
\newblock
\urldef\tempurl%
\url{https://cogsci.mindmodeling.org/2020/papers/0618/index.html}
\showURL{%
\tempurl}


\bibitem[Ma et~al\mbox{.}(2021)]%
        {Ma2021MixedModalityII}
\bibfield{author}{\bibinfo{person}{Yuan Ma}, \bibinfo{person}{Timm Kleemann}, {and} \bibinfo{person}{J{\"{u}}rgen Ziegler}.} \bibinfo{year}{2021}\natexlab{}.
\newblock \showarticletitle{Mixed-Modality Interaction in Conversational Recommender Systems}. In \bibinfo{booktitle}{\emph{Proceedings of the 8th Joint Workshop on Interfaces and Human Decision Making for Recommender Systems co-located with 15th {ACM} Conference on Recommender Systems (RecSys 2021), Online Event, September 25 and September 29, 2021}} \emph{(\bibinfo{series}{{CEUR} Workshop Proceedings}, Vol.~\bibinfo{volume}{2948})}, \bibfield{editor}{\bibinfo{person}{Peter Brusilovsky}, \bibinfo{person}{Marco de~Gemmis}, \bibinfo{person}{Alexander Felfernig}, \bibinfo{person}{Elisabeth Lex}, \bibinfo{person}{Pasquale Lops}, \bibinfo{person}{Giovanni Semeraro}, {and} \bibinfo{person}{Martijn~C. Willemsen}} (Eds.). \bibinfo{publisher}{CEUR-WS.org}, \bibinfo{pages}{21--37}.
\newblock
\urldef\tempurl%
\url{https://ceur-ws.org/Vol-2948/paper2.pdf}
\showURL{%
\tempurl}


\bibitem[Mayer and Moreno(2003)]%
        {mayer2003nine}
\bibfield{author}{\bibinfo{person}{Richard~E. Mayer} {and} \bibinfo{person}{Roxana Moreno}.} \bibinfo{year}{2003}\natexlab{}.
\newblock \showarticletitle{Nine Ways to Reduce Cognitive Load in Multimedia Learning}.
\newblock \bibinfo{journal}{\emph{Educational Psychologist}} \bibinfo{volume}{38}, \bibinfo{number}{1} (\bibinfo{year}{2003}), \bibinfo{pages}{43--52}.
\newblock


\bibitem[Meng et~al\mbox{.}(2021)]%
        {Meng2021InitiativeAwareSL}
\bibfield{author}{\bibinfo{person}{Chuan Meng}, \bibinfo{person}{Pengjie Ren}, \bibinfo{person}{Zhumin Chen}, \bibinfo{person}{Zhaochun Ren}, \bibinfo{person}{Tengxiao Xi}, {and} \bibinfo{person}{Maarten de Rijke}.} \bibinfo{year}{2021}\natexlab{}.
\newblock \showarticletitle{Initiative-Aware Self-Supervised Learning for Knowledge-Grounded Conversations}. In \bibinfo{booktitle}{\emph{{SIGIR} '21: The 44th International {ACM} {SIGIR} Conference on Research and Development in Information Retrieval, Virtual Event, Canada, July 11-15, 2021}}, \bibfield{editor}{\bibinfo{person}{Fernando Diaz}, \bibinfo{person}{Chirag Shah}, \bibinfo{person}{Torsten Suel}, \bibinfo{person}{Pablo Castells}, \bibinfo{person}{Rosie Jones}, {and} \bibinfo{person}{Tetsuya Sakai}} (Eds.). \bibinfo{publisher}{{ACM}}, \bibinfo{pages}{522--532}.
\newblock
\urldef\tempurl%
\url{https://doi.org/10.1145/3404835.3462824}
\showDOI{\tempurl}


\bibitem[Murrugarra{-}Llerena and Kovashka(2018)]%
        {MurrugarraLlerena2018ImageRW}
\bibfield{author}{\bibinfo{person}{Nils Murrugarra{-}Llerena} {and} \bibinfo{person}{Adriana Kovashka}.} \bibinfo{year}{2018}\natexlab{}.
\newblock \showarticletitle{Image Retrieval with Mixed Initiative and Multimodal Feedback}. In \bibinfo{booktitle}{\emph{British Machine Vision Conference 2018, {BMVC} 2018, Newcastle, UK, September 3-6, 2018}}. \bibinfo{publisher}{{BMVA} Press}, \bibinfo{pages}{310}.
\newblock
\urldef\tempurl%
\url{http://bmvc2018.org/contents/papers/0151.pdf}
\showURL{%
\tempurl}


\bibitem[Nowak and R{\"{u}}ger(2010)]%
        {NowakR10-reliable-annotations}
\bibfield{author}{\bibinfo{person}{Stefanie Nowak} {and} \bibinfo{person}{Stefan~M. R{\"{u}}ger}.} \bibinfo{year}{2010}\natexlab{}.
\newblock \showarticletitle{How Reliable are Annotations via Crowdsourcing: A Study about Inter-annotator Agreement for Multi-label Image Annotation}. In \bibinfo{booktitle}{\emph{Proceedings of the 11th {Association for Computing Machinery} {SIGMM} International Conference on Multimedia Information Retrieval, {MIR} 2010, Philadelphia, Pennsylvania, USA, March 29-31, 2010}}, \bibfield{editor}{\bibinfo{person}{James~Ze Wang}, \bibinfo{person}{Nozha Boujemaa}, \bibinfo{person}{Nuria~Oliver Ramirez}, {and} \bibinfo{person}{Apostol Natsev}} (Eds.). \bibinfo{publisher}{{Association for Computing Machinery}}, \bibinfo{pages}{557--566}.
\newblock
\urldef\tempurl%
\url{https://doi.org/10.1145/1743384.1743478}
\showDOI{\tempurl}


\bibitem[O'Brien et~al\mbox{.}(2022)]%
        {DBLP:conf/chiir/OBrienCKB22}
\bibfield{author}{\bibinfo{person}{Heather O'Brien}, \bibinfo{person}{Amelia~W. Cole}, \bibinfo{person}{Andrea Kampen}, {and} \bibinfo{person}{Kathy Brennan}.} \bibinfo{year}{2022}\natexlab{}.
\newblock \showarticletitle{The Effects of Domain and Search Expertise on Learning Outcomes in Digital Library Use}. In \bibinfo{booktitle}{\emph{{CHIIR} '22: {ACM} {SIGIR} Conference on Human Information Interaction and Retrieval, Regensburg, Germany, March 14 - 18, 2022}}, \bibfield{editor}{\bibinfo{person}{David Elsweiler}} (Ed.). \bibinfo{publisher}{{ACM}}, \bibinfo{pages}{202--210}.
\newblock
\urldef\tempurl%
\url{https://doi.org/10.1145/3498366.3505761}
\showDOI{\tempurl}


\bibitem[Owoicho et~al\mbox{.}(2023)]%
        {Owoicho2023ExploitingSU}
\bibfield{author}{\bibinfo{person}{Paul Owoicho}, \bibinfo{person}{Ivan Sekulic}, \bibinfo{person}{Mohammad Aliannejadi}, \bibinfo{person}{Jeffrey Dalton}, {and} \bibinfo{person}{Fabio Crestani}.} \bibinfo{year}{2023}\natexlab{}.
\newblock \showarticletitle{Exploiting Simulated User Feedback for Conversational Search: Ranking, Rewriting, and Beyond}.
\newblock  (\bibinfo{year}{2023}), \bibinfo{pages}{632--642}.
\newblock
\urldef\tempurl%
\url{https://doi.org/10.1145/3539618.3591683}
\showDOI{\tempurl}


\bibitem[Paivio(1991)]%
        {paivio1991dual}
\bibfield{author}{\bibinfo{person}{Allan Paivio}.} \bibinfo{year}{1991}\natexlab{}.
\newblock \showarticletitle{Dual Coding Theory: Retrospect and Current Status}.
\newblock \bibinfo{journal}{\emph{Canadian Journal of Psychology/Revue Canadienne de Psychologie}} \bibinfo{volume}{45}, \bibinfo{number}{3} (\bibinfo{year}{1991}), \bibinfo{pages}{255}.
\newblock


\bibitem[Qu et~al\mbox{.}(2018)]%
        {DBLP:conf/sigir/QuYCTZQ18}
\bibfield{author}{\bibinfo{person}{Chen Qu}, \bibinfo{person}{Liu Yang}, \bibinfo{person}{W.~Bruce Croft}, \bibinfo{person}{Johanne~R. Trippas}, \bibinfo{person}{Yongfeng Zhang}, {and} \bibinfo{person}{Minghui Qiu}.} \bibinfo{year}{2018}\natexlab{}.
\newblock \showarticletitle{Analyzing and Characterizing User Intent in Information-seeking Conversations}. In \bibinfo{booktitle}{\emph{The 41st International {ACM} {SIGIR} Conference on Research {\&} Development in Information Retrieval, {SIGIR} 2018, Ann Arbor, MI, USA, July 08-12, 2018}}, \bibfield{editor}{\bibinfo{person}{Kevyn Collins{-}Thompson}, \bibinfo{person}{Qiaozhu Mei}, \bibinfo{person}{Brian~D. Davison}, \bibinfo{person}{Yiqun Liu}, {and} \bibinfo{person}{Emine Yilmaz}} (Eds.). \bibinfo{publisher}{{ACM}}, \bibinfo{pages}{989--992}.
\newblock
\urldef\tempurl%
\url{https://doi.org/10.1145/3209978.3210124}
\showDOI{\tempurl}


\bibitem[Radford et~al\mbox{.}(2021)]%
        {Radford2021LearningTV}
\bibfield{author}{\bibinfo{person}{Alec Radford}, \bibinfo{person}{Jong~Wook Kim}, \bibinfo{person}{Chris Hallacy}, \bibinfo{person}{Aditya Ramesh}, \bibinfo{person}{Gabriel Goh}, \bibinfo{person}{Sandhini Agarwal}, \bibinfo{person}{Girish Sastry}, \bibinfo{person}{Amanda Askell}, \bibinfo{person}{Pamela Mishkin}, \bibinfo{person}{Jack Clark}, \bibinfo{person}{Gretchen Krueger}, {and} \bibinfo{person}{Ilya Sutskever}.} \bibinfo{year}{2021}\natexlab{}.
\newblock \showarticletitle{Learning Transferable Visual Models From Natural Language Supervision}. In \bibinfo{booktitle}{\emph{Proceedings of the 38th International Conference on Machine Learning, {ICML} 2021, 18-24 July 2021, Virtual Event}} \emph{(\bibinfo{series}{Proceedings of Machine Learning Research}, Vol.~\bibinfo{volume}{139})}, \bibfield{editor}{\bibinfo{person}{Marina Meila} {and} \bibinfo{person}{Tong Zhang}} (Eds.). \bibinfo{publisher}{{PMLR}}, \bibinfo{pages}{8748--8763}.
\newblock
\urldef\tempurl%
\url{http://proceedings.mlr.press/v139/radford21a.html}
\showURL{%
\tempurl}


\bibitem[Radlinski and Craswell(2017)]%
        {Radlinski2017Theoretical}
\bibfield{author}{\bibinfo{person}{Filip Radlinski} {and} \bibinfo{person}{Nick Craswell}.} \bibinfo{year}{2017}\natexlab{}.
\newblock \showarticletitle{A Theoretical Framework for Conversational Search}. In \bibinfo{booktitle}{\emph{Proceedings of the 2017 Conference on Conference Human Information Interaction and Retrieval, {CHIIR} 2017, Oslo, Norway, March 7-11, 2017}}. \bibinfo{publisher}{{ACM}}, \bibinfo{pages}{117--126}.
\newblock
\urldef\tempurl%
\url{https://doi.org/10.1145/3020165.3020183}
\showDOI{\tempurl}


\bibitem[Rao and III(2018)]%
        {Rao2018LearningTA}
\bibfield{author}{\bibinfo{person}{Sudha Rao} {and} \bibinfo{person}{Hal~Daum{\'{e}} III}.} \bibinfo{year}{2018}\natexlab{}.
\newblock \showarticletitle{Learning to Ask Good Questions: Ranking Clarification Questions using Neural Expected Value of Perfect Information}. In \bibinfo{booktitle}{\emph{Proceedings of the 56th Annual Meeting of the Association for Computational Linguistics, {ACL} 2018, Melbourne, Australia, July 15-20, 2018, Volume 1: Long Papers}}, \bibfield{editor}{\bibinfo{person}{Iryna Gurevych} {and} \bibinfo{person}{Yusuke Miyao}} (Eds.). \bibinfo{publisher}{Association for Computational Linguistics}, \bibinfo{pages}{2737--2746}.
\newblock
\urldef\tempurl%
\url{https://doi.org/10.18653/V1/P18-1255}
\showDOI{\tempurl}


\bibitem[Sekulic et~al\mbox{.}(2021a)]%
        {Sekulic2021UserEP}
\bibfield{author}{\bibinfo{person}{Ivan Sekulic}, \bibinfo{person}{Mohammad Aliannejadi}, {and} \bibinfo{person}{Fabio Crestani}.} \bibinfo{year}{2021}\natexlab{a}.
\newblock \showarticletitle{User Engagement Prediction for Clarification in Search}. In \bibinfo{booktitle}{\emph{Advances in Information Retrieval - 43rd European Conference on {IR} Research, {ECIR} 2021, Virtual Event, March 28 - April 1, 2021, Proceedings, Part {I}}} \emph{(\bibinfo{series}{Lecture Notes in Computer Science}, Vol.~\bibinfo{volume}{12656})}, \bibfield{editor}{\bibinfo{person}{Djoerd Hiemstra}, \bibinfo{person}{Marie{-}Francine Moens}, \bibinfo{person}{Josiane Mothe}, \bibinfo{person}{Raffaele Perego}, \bibinfo{person}{Martin Potthast}, {and} \bibinfo{person}{Fabrizio Sebastiani}} (Eds.). \bibinfo{publisher}{Springer}, \bibinfo{pages}{619--633}.
\newblock
\urldef\tempurl%
\url{https://doi.org/10.1007/978-3-030-72113-8\_41}
\showDOI{\tempurl}


\bibitem[Sekulic et~al\mbox{.}(2021b)]%
        {Sekulic2021User}
\bibfield{author}{\bibinfo{person}{Ivan Sekulic}, \bibinfo{person}{Mohammad Aliannejadi}, {and} \bibinfo{person}{Fabio Crestani}.} \bibinfo{year}{2021}\natexlab{b}.
\newblock \showarticletitle{User Engagement Prediction for Clarification in Search}. In \bibinfo{booktitle}{\emph{{ECIR} {(1)}}} \emph{(\bibinfo{series}{Lecture Notes in Computer Science}, Vol.~\bibinfo{volume}{12656})}. \bibinfo{publisher}{Springer}, \bibinfo{pages}{619--633}.
\newblock


\bibitem[Siro et~al\mbox{.}(2024)]%
        {siro-agentcq}
\bibfield{author}{\bibinfo{person}{Clemencia Siro}, \bibinfo{person}{Yifei Yuan}, \bibinfo{person}{Mohammad Aliannejadi}, {and} \bibinfo{person}{Maarten de Rijke}.} \bibinfo{year}{2024}\natexlab{}.
\newblock \showarticletitle{{AGENT-CQ:} Automatic Generation and Evaluation of Clarifying Questions for Conversational Search with LLMs}.
\newblock \bibinfo{journal}{\emph{CoRR}}  \bibinfo{volume}{abs/2410.19692} (\bibinfo{year}{2024}).
\newblock


\bibitem[Talmor et~al\mbox{.}(2021)]%
        {Talmor2021MultiModalQACQ}
\bibfield{author}{\bibinfo{person}{Alon Talmor}, \bibinfo{person}{Ori Yoran}, \bibinfo{person}{Amnon Catav}, \bibinfo{person}{Dan Lahav}, \bibinfo{person}{Yizhong Wang}, \bibinfo{person}{Akari Asai}, \bibinfo{person}{Gabriel Ilharco}, \bibinfo{person}{Hannaneh Hajishirzi}, {and} \bibinfo{person}{Jonathan Berant}.} \bibinfo{year}{2021}\natexlab{}.
\newblock \showarticletitle{MultiModalQA: Complex Question Answering over Text, Tables and Images}.
\newblock  (\bibinfo{year}{2021}).
\newblock
\urldef\tempurl%
\url{https://openreview.net/forum?id=ee6W5UgQLa}
\showURL{%
\tempurl}


\bibitem[Turk(2014)]%
        {turk-2014-multimodal}
\bibfield{author}{\bibinfo{person}{Matthew~A. Turk}.} \bibinfo{year}{2014}\natexlab{}.
\newblock \showarticletitle{Multimodal Interaction: {A} Review}.
\newblock \bibinfo{journal}{\emph{Pattern Recognit. Lett.}}  \bibinfo{volume}{36} (\bibinfo{year}{2014}), \bibinfo{pages}{189--195}.
\newblock
\urldef\tempurl%
\url{https://doi.org/10.1016/J.PATREC.2013.07.003}
\showDOI{\tempurl}


\bibitem[Vakulenko et~al\mbox{.}(2020)]%
        {Vakulenko2020AnAO}
\bibfield{author}{\bibinfo{person}{Svitlana Vakulenko}, \bibinfo{person}{Evangelos Kanoulas}, {and} \bibinfo{person}{Maarten de Rijke}.} \bibinfo{year}{2020}\natexlab{}.
\newblock \showarticletitle{An Analysis of Mixed Initiative and Collaboration in Information-Seeking Dialogues}. In \bibinfo{booktitle}{\emph{Proceedings of the 43rd International {ACM} {SIGIR} conference on research and development in Information Retrieval, {SIGIR} 2020, Virtual Event, China, July 25-30, 2020}}, \bibfield{editor}{\bibinfo{person}{Jimmy~X. Huang}, \bibinfo{person}{Yi~Chang}, \bibinfo{person}{Xueqi Cheng}, \bibinfo{person}{Jaap Kamps}, \bibinfo{person}{Vanessa Murdock}, \bibinfo{person}{Ji{-}Rong Wen}, {and} \bibinfo{person}{Yiqun Liu}} (Eds.). \bibinfo{publisher}{{ACM}}, \bibinfo{pages}{2085--2088}.
\newblock
\urldef\tempurl%
\url{https://doi.org/10.1145/3397271.3401297}
\showDOI{\tempurl}


\bibitem[Wang et~al\mbox{.}(2024)]%
        {Wang2024What}
\bibfield{author}{\bibinfo{person}{Silang Wang}, \bibinfo{person}{Hyeongcheol Kim}, \bibinfo{person}{Nuwan Janaka}, \bibinfo{person}{Kun Yue}, \bibinfo{person}{Hoang{-}Long Nguyen}, \bibinfo{person}{Shengdong Zhao}, \bibinfo{person}{Haiming Liu}, {and} \bibinfo{person}{Khanh{-}Duy Le}.} \bibinfo{year}{2024}\natexlab{}.
\newblock \showarticletitle{"What's this?": Understanding User Interaction Behaviour with Multimodal Input Information Retrieval System}. In \bibinfo{booktitle}{\emph{MobileHCI (Companion)}}. \bibinfo{publisher}{{ACM}}, \bibinfo{pages}{3:1--3:7}.
\newblock


\bibitem[Wang et~al\mbox{.}(2023)]%
        {Wang2023GenerativeQR}
\bibfield{author}{\bibinfo{person}{Xiao Wang}, \bibinfo{person}{Sean MacAvaney}, \bibinfo{person}{Craig Macdonald}, {and} \bibinfo{person}{Iadh Ounis}.} \bibinfo{year}{2023}\natexlab{}.
\newblock \showarticletitle{Generative Query Reformulation for Effective Adhoc Search}.
\newblock \bibinfo{journal}{\emph{CoRR}}  \bibinfo{volume}{abs/2308.00415} (\bibinfo{year}{2023}).
\newblock
\urldef\tempurl%
\url{https://doi.org/10.48550/ARXIV.2308.00415}
\showDOI{\tempurl}
\showeprint[arXiv]{2308.00415}


\bibitem[Wang and Ai(2021)]%
        {Wang2021Controlling}
\bibfield{author}{\bibinfo{person}{Zhenduo Wang} {and} \bibinfo{person}{Qingyao Ai}.} \bibinfo{year}{2021}\natexlab{}.
\newblock \showarticletitle{Controlling the Risk of Conversational Search via Reinforcement Learning}. In \bibinfo{booktitle}{\emph{{WWW}}}. \bibinfo{publisher}{{ACM} / {IW3C2}}, \bibinfo{pages}{1968--1977}.
\newblock


\bibitem[White et~al\mbox{.}(2009)]%
        {DBLP:conf/wsdm/WhiteDT09}
\bibfield{author}{\bibinfo{person}{Ryen~W. White}, \bibinfo{person}{Susan~T. Dumais}, {and} \bibinfo{person}{Jaime Teevan}.} \bibinfo{year}{2009}\natexlab{}.
\newblock \showarticletitle{Characterizing the Influence of Domain Expertise on Web Search Behavior}. In \bibinfo{booktitle}{\emph{Proceedings of the Second International Conference on Web Search and Web Data Mining, {WSDM} 2009, Barcelona, Spain, February 9-11, 2009}}, \bibfield{editor}{\bibinfo{person}{Ricardo Baeza{-}Yates}, \bibinfo{person}{Paolo Boldi}, \bibinfo{person}{Berthier~A. Ribeiro{-}Neto}, {and} \bibinfo{person}{Berkant~Barla Cambazoglu}} (Eds.). \bibinfo{publisher}{{ACM}}, \bibinfo{pages}{132--141}.
\newblock
\urldef\tempurl%
\url{https://doi.org/10.1145/1498759.1498819}
\showDOI{\tempurl}


\bibitem[Xu et~al\mbox{.}(2019)]%
        {Xu2019AskingCQ}
\bibfield{author}{\bibinfo{person}{Jingjing Xu}, \bibinfo{person}{Yuechen Wang}, \bibinfo{person}{Duyu Tang}, \bibinfo{person}{Nan Duan}, \bibinfo{person}{Pengcheng Yang}, \bibinfo{person}{Qi Zeng}, \bibinfo{person}{Ming Zhou}, {and} \bibinfo{person}{Xu Sun}.} \bibinfo{year}{2019}\natexlab{}.
\newblock \showarticletitle{Asking Clarification Questions in Knowledge-Based Question Answering}. In \bibinfo{booktitle}{\emph{Proceedings of the 2019 Conference on Empirical Methods in Natural Language Processing and the 9th International Joint Conference on Natural Language Processing, {EMNLP-IJCNLP} 2019, Hong Kong, China, November 3-7, 2019}}, \bibfield{editor}{\bibinfo{person}{Kentaro Inui}, \bibinfo{person}{Jing Jiang}, \bibinfo{person}{Vincent Ng}, {and} \bibinfo{person}{Xiaojun Wan}} (Eds.). \bibinfo{publisher}{Association for Computational Linguistics}, \bibinfo{pages}{1618--1629}.
\newblock
\urldef\tempurl%
\url{https://doi.org/10.18653/V1/D19-1172}
\showDOI{\tempurl}


\bibitem[Yuan et~al\mbox{.}(2022)]%
        {Yuan2022McQueenAB}
\bibfield{author}{\bibinfo{person}{Yifei Yuan}, \bibinfo{person}{Chen Shi}, \bibinfo{person}{Runze Wang}, \bibinfo{person}{Liyi Chen}, \bibinfo{person}{Feijun Jiang}, \bibinfo{person}{Yuan You}, {and} \bibinfo{person}{Wai Lam}.} \bibinfo{year}{2022}\natexlab{}.
\newblock \showarticletitle{McQueen: a Benchmark for Multimodal Conversational Query Rewrite}. In \bibinfo{booktitle}{\emph{Proceedings of the 2022 Conference on Empirical Methods in Natural Language Processing, {EMNLP} 2022, Abu Dhabi, United Arab Emirates, December 7-11, 2022}}, \bibfield{editor}{\bibinfo{person}{Yoav Goldberg}, \bibinfo{person}{Zornitsa Kozareva}, {and} \bibinfo{person}{Yue Zhang}} (Eds.). \bibinfo{publisher}{Association for Computational Linguistics}, \bibinfo{pages}{4834--4844}.
\newblock
\urldef\tempurl%
\url{https://doi.org/10.18653/V1/2022.EMNLP-MAIN.320}
\showDOI{\tempurl}


\bibitem[Yuan et~al\mbox{.}(2024)]%
        {Yuan2024Asking}
\bibfield{author}{\bibinfo{person}{Yifei Yuan}, \bibinfo{person}{Clemencia Siro}, \bibinfo{person}{Mohammad Aliannejadi}, \bibinfo{person}{Maarten de Rijke}, {and} \bibinfo{person}{Wai Lam}.} \bibinfo{year}{2024}\natexlab{}.
\newblock \showarticletitle{Asking Multimodal Clarifying Questions in Mixed-Initiative Conversational Search}. In \bibinfo{booktitle}{\emph{Proceedings of the ACM Web Conference 2024}}. \bibinfo{publisher}{{ACM}}, \bibinfo{pages}{1474--1485}.
\newblock


\bibitem[Zamani et~al\mbox{.}(2020a)]%
        {Zamani2020GeneratingCQ}
\bibfield{author}{\bibinfo{person}{Hamed Zamani}, \bibinfo{person}{Susan~T. Dumais}, \bibinfo{person}{Nick Craswell}, \bibinfo{person}{Paul~N. Bennett}, {and} \bibinfo{person}{Gord Lueck}.} \bibinfo{year}{2020}\natexlab{a}.
\newblock \showarticletitle{Generating Clarifying Questions for Information Retrieval}. In \bibinfo{booktitle}{\emph{{WWW} '20: The Web Conference 2020, Taipei, Taiwan, April 20-24, 2020}}, \bibfield{editor}{\bibinfo{person}{Yennun Huang}, \bibinfo{person}{Irwin King}, \bibinfo{person}{Tie{-}Yan Liu}, {and} \bibinfo{person}{Maarten van Steen}} (Eds.). \bibinfo{publisher}{{ACM} / {IW3C2}}, \bibinfo{pages}{418--428}.
\newblock
\urldef\tempurl%
\url{https://doi.org/10.1145/3366423.3380126}
\showDOI{\tempurl}


\bibitem[Zamani et~al\mbox{.}(2020b)]%
        {Zamani2020MIMICSAL}
\bibfield{author}{\bibinfo{person}{Hamed Zamani}, \bibinfo{person}{Gord Lueck}, \bibinfo{person}{Everest Chen}, \bibinfo{person}{Rodolfo Quispe}, \bibinfo{person}{Flint Luu}, {and} \bibinfo{person}{Nick Craswell}.} \bibinfo{year}{2020}\natexlab{b}.
\newblock \showarticletitle{{MIMICS:} {A} Large-Scale Data Collection for Search Clarification}. In \bibinfo{booktitle}{\emph{{CIKM} '20: The 29th {ACM} International Conference on Information and Knowledge Management, Virtual Event, Ireland, October 19-23, 2020}}, \bibfield{editor}{\bibinfo{person}{Mathieu d'Aquin}, \bibinfo{person}{Stefan Dietze}, \bibinfo{person}{Claudia Hauff}, \bibinfo{person}{Edward Curry}, {and} \bibinfo{person}{Philippe Cudr{\'{e}}{-}Mauroux}} (Eds.). \bibinfo{publisher}{{ACM}}, \bibinfo{pages}{3189--3196}.
\newblock
\urldef\tempurl%
\url{https://doi.org/10.1145/3340531.3412772}
\showDOI{\tempurl}


\bibitem[Zamani et~al\mbox{.}(2020c)]%
        {Zamani2020AnalyzingAL}
\bibfield{author}{\bibinfo{person}{Hamed Zamani}, \bibinfo{person}{Bhaskar Mitra}, \bibinfo{person}{Everest Chen}, \bibinfo{person}{Gord Lueck}, \bibinfo{person}{Fernando Diaz}, \bibinfo{person}{Paul~N. Bennett}, \bibinfo{person}{Nick Craswell}, {and} \bibinfo{person}{Susan~T. Dumais}.} \bibinfo{year}{2020}\natexlab{c}.
\newblock \showarticletitle{Analyzing and Learning from User Interactions for Search Clarification}. In \bibinfo{booktitle}{\emph{Proceedings of the 43rd International {ACM} {SIGIR} conference on research and development in Information Retrieval, {SIGIR} 2020, Virtual Event, China, July 25-30, 2020}}, \bibfield{editor}{\bibinfo{person}{Jimmy~X. Huang}, \bibinfo{person}{Yi~Chang}, \bibinfo{person}{Xueqi Cheng}, \bibinfo{person}{Jaap Kamps}, \bibinfo{person}{Vanessa Murdock}, \bibinfo{person}{Ji{-}Rong Wen}, {and} \bibinfo{person}{Yiqun Liu}} (Eds.). \bibinfo{publisher}{{ACM}}, \bibinfo{pages}{1181--1190}.
\newblock
\urldef\tempurl%
\url{https://doi.org/10.1145/3397271.3401160}
\showDOI{\tempurl}


\bibitem[Zou et~al\mbox{.}(2023)]%
        {Zou2023Users}
\bibfield{author}{\bibinfo{person}{Jie Zou}, \bibinfo{person}{Mohammad Aliannejadi}, \bibinfo{person}{Evangelos Kanoulas}, \bibinfo{person}{Maria~Soledad Pera}, {and} \bibinfo{person}{Yiqun Liu}.} \bibinfo{year}{2023}\natexlab{}.
\newblock \showarticletitle{Users Meet Clarifying Questions: Toward a Better Understanding of User Interactions for Search Clarification}.
\newblock \bibinfo{journal}{\emph{{ACM} Trans. Inf. Syst.}} \bibinfo{volume}{41}, \bibinfo{number}{1} (\bibinfo{year}{2023}), \bibinfo{pages}{16:1--16:25}.
\newblock


\bibitem[Zou and Kanoulas(2019)]%
        {Zou2019Learning}
\bibfield{author}{\bibinfo{person}{Jie Zou} {and} \bibinfo{person}{Evangelos Kanoulas}.} \bibinfo{year}{2019}\natexlab{}.
\newblock \showarticletitle{Learning to Ask: Question-based Sequential Bayesian Product Search}. In \bibinfo{booktitle}{\emph{Proceedings of the 28th ACM International Conference on Information and Knowledge Management}}. \bibinfo{publisher}{{ACM}}, \bibinfo{pages}{369--378}.
\newblock


\end{thebibliography}

\end{document}